\documentclass[journal]{IEEEtran}
\usepackage{cite}
\usepackage{amsmath,amssymb,amsfonts}
\usepackage{algorithm,algorithmic}%
\usepackage{graphicx}
\usepackage{textcomp}
\usepackage{multirow}
\usepackage{array,booktabs}
\usepackage{subcaption}
\usepackage{url} 
\usepackage{hyperref}
\usepackage{soul}
\usepackage{amsmath,amssymb,amsfonts}
\usepackage{arydshln}

\usepackage{pifont}
\newcommand{\xmark}{\ding{55}}%

\soulregister{\ref}{7} 
\soulregister{\cite}{7} 

\usepackage{bm}
\usepackage{fontawesome5}

\hypersetup{
    colorlinks=true,
    citecolor=blue
}

\begin{document}
\title{\includegraphics[height=1.5ex]{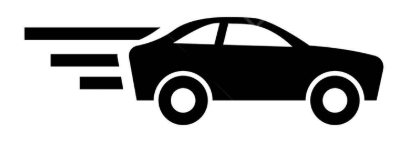} \textbf{TriLiteNet}\\
Lightweight Model for Multi-Task Visual Perception}


\author{\IEEEauthorblockN{Quang-Huy Che\IEEEauthorrefmark{1}\IEEEauthorrefmark{3}, Duc-Khai Lam\IEEEauthorrefmark{2}\IEEEauthorrefmark{3}}

\IEEEauthorblockA{\IEEEauthorrefmark{1}Laboratory of Multimedia Communications, 
University of Information Technology, Ho Chi Minh City, Vietnam}

\IEEEauthorblockA{\IEEEauthorrefmark{2}Faculty of Computer Engineering, University of Information Technology, Ho Chi Minh City, Vietnam}

\IEEEauthorblockA{\IEEEauthorrefmark{3}Vietnam National University, Ho Chi Minh City, Vietnam \\
Email: \texttt{huycq@uit.edu.vn,khaild@uit.edu.vn}}}

\maketitle

\begin{abstract}
Efficient perception models are essential for Advanced Driver Assistance Systems (ADAS), as these applications require rapid processing and response to ensure safety and effectiveness in real-world environments. To address the real-time execution needs of such perception models, this study introduces the TriLiteNet model. This model can simultaneously manage multiple tasks related to panoramic driving perception. TriLiteNet is designed to optimize performance while maintaining low computational costs. Experimental results on the BDD100k dataset demonstrate that the model achieves competitive performance across three key tasks: vehicle detection, drivable area segmentation, and lane line segmentation. Specifically, the TriLiteNet$_{base}$ demonstrated a recall of 85.6\% for vehicle detection, a mean Intersection over Union (mIoU) of 92.4\% for drivable area segmentation, and an Acc of 82.3\% for lane line segmentation with only 2.35M parameters and a computational cost of 7.72 GFLOPs. Our proposed model includes a \textit{tiny} configuration with just 0.14M parameters, which provides a multi-task solution with minimal computational demand. Evaluated for latency and power consumption on embedded devices, TriLiteNet in both configurations shows low latency and reasonable power during inference. By balancing performance, computational efficiency, and scalability, TriLiteNet offers a practical and deployable solution for real-world autonomous driving applications. Code is available at \url{https://github.com/chequanghuy/TriLiteNet}.
\end{abstract}


\maketitle

\section{Introduction}
\label{sec:introduction}
Efficient perception models play a vital role in enhancing the performance of Advanced Driver Assistance Systems (ADAS) in environments with limited resources. Initially, these systems depended on traditional sensor technologies like radar and LiDAR to understand their surroundings. The early systems utilized rule-based algorithms and manual features to process the sensor data, which allowed for essential object recognition and navigation capabilities. However, these initial approaches had limitations in accuracy and adaptability, especially in complex and dynamic driving environments. With advancements in deep learning, camera-based perception has emerged as a focal point, with panoramic driving perception as an effective solution for autonomous vehicles. This approach captures comprehensive semantic information from the environment, providing the necessary data foundation to optimize driving decisions. \cite{yolov5, yolov8, faster, Sparse-R-CNN,GCNet, pspnet, dnlnet,enet, enet-sad} have proposed three primary tasks in panoramic driving perception: vehicle detection, drivable area segmentation, and lane line segmentation. 

\begin{itemize}
    \item \textbf{Vehicle detection} identifies the coordinates of traffic objects, assisting the vehicle in predicting and reacting to potential collision risks.
    \item \textbf{Drivable area segmentation} defines regions where vehicles can operate safely, enabling precise control predictions.
    \item \textbf{Lane segmentation} provides detailed information about lanes, supporting the vehicle in maintaining or changing lanes securely.
\end{itemize}

When these three tasks are integrated, autonomous vehicles can comprehensively understand their environment, enabling more accurate and efficient decision-making. Many deep learning models, particularly Convolutional Neural Networks (CNNs), have significantly succeeded in panoramic driving perception tasks. Single-task models have demonstrated promising performance in various applications: \cite{yolov5, yolov8, faster, Sparse-R-CNN} for vehicle detection, \cite{GCNet, pspnet, dnlnet} for drivable area segmentation, and \cite{enet, enet-sad} for lane segmentation. However, these single-task models are limited to handling only one task at a time, necessitating multiple models to obtain outputs for different tasks. This approach significantly increases computational costs when performing inference with multiple models simultaneously, making them unsuitable for autonomous driving systems with constrained computational resources.

Multi-task models \cite{yolop, yolopv2, yolopv3, YOLOPX} have emerged as an efficient solution, enabling information sharing across tasks to reduce computational costs and improve generalization capabilities. Results on the challenging BDD100K dataset \cite{bdd100k} demonstrate that training these tasks simultaneously delivers notable accuracy and computational efficiency performance. However, these models, with parameters ranging from a few million \cite{yolop, ayolom} to tens of millions \cite{yolopv2, yolopv3, YOLOPX, hybridnets}, often rely on complex architectures, making it difficult to achieve real-time performance on low-cost embedded devices. This highlights the need to optimize multi-task architectures for resource-constrained systems, enhancing their practical applicability in autonomous driving. Che et al. \cite{twin, twinplus} have attempted to reduce the computational cost of these models by introducing architectures capable of simultaneously performing two segmentation tasks: drivable area segmentation and lane line segmentation. These models leverage dilated convolutional layers with varying dilation rates and achieve competitive results with fewer than one million parameters compared to previous state-of-the-art (SOTA) models. This is one of the few proposals explicitly targeting low computational cost. However, this approach focuses only on segmentation tasks, neglecting object detection—a critical component in autonomous driving systems. Therefore, developing multi-task models that optimize performance and computational cost remains a crucial research direction to unlock their potential for real-world applications.

In this study, we propose a lightweight multi-task model where a shared encoder processes the input image to extract features. These features are pre-processed and distributed to three separate decoders for different tasks: vehicle detection, drivable area segmentation, and lane line segmentation.. The main contributions of this research are summarized as follows:

\begin{figure*}
    \includegraphics[width=\textwidth]{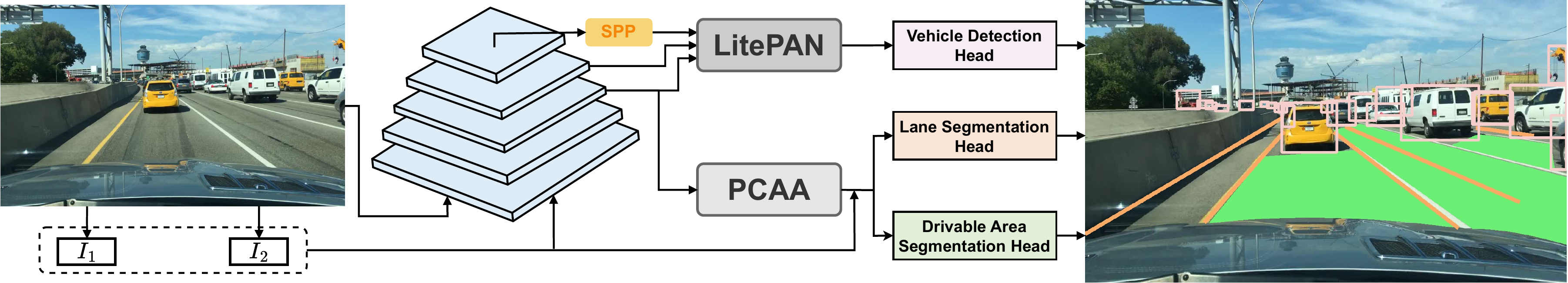}
    \caption{TriLiteNet is a unified encoder-decoder network consisting of a shared encoder and three separate decoders for different tasks: vehicle detection, drivable area segmentation, and lane line segmentation. In the encoding stage, the TwinLiteNet$^+$ encoder block is adapted to extract features that are subsequently directed to distinct decoder blocks. For segmentation tasks, the two segmentation heads are designed similarly, taking the output of the PCAA block as input. In contrast, the decoder for vehicle detection includes LitePAN and SPP blocks to enhance feature extraction, particularly for small and complex objects. This process generates three feature maps of different scales, which are then fed into the object detection head for improved detection performance. Additionally, $I_1$ and $I_2$ are lower-resolution input images obtained through average pooling operations, which effectively downsample the input image while preserving essential contextual information. These downsampled images are concatenated with feature maps during the encoding and decoding of the segmentation tasks, enhancing the information flow across different resolution levels.}
    \label{fig:pipeline}
\end{figure*}

\begin{itemize}

    \item We introduce TriLiteNet, a lightweight multi-task model optimized for low computational cost. The encoder leverages Efficient Spatial Pyramid (ESP) blocks using depth-wise separable convolutions, while the two decoders for segmentation tasks utilize simple upsampling stages based on transposed and standard convolutions. Our approach simplifies the traditional complex neck and head designs for the vehicle detection task by introducing LitePAN, inspired by the Path Aggregation Network (PAN) architecture, which incorporates depth-wise separable convolutions and a streamlined detection head. 

    \item The TriLiteNet model is developed in three configurations to cater to diverse resource and performance requirements. The largest configuration, TriLiteNet$_{base}$, consists of 2.35M parameters and requires 7.72 GFLOPs. Additionally, two compact versions—TriLiteNet$_{small}$ with 0.59M parameters and 1.99 GFLOPs, and TriLiteNet$_{tiny}$ with only 0.15M parameters and 0.55 GFLOPs—are optimized for deployment on resource-constrained systems.

    \item Experimental evaluation on the BDD100K dataset demonstrates the competitive performance of TriLiteNet$_{base}$, achieving an 85.6\% recall for vehicle detection, a mean Intersection over Union (mIoU) of 92.4\% for drivable area segmentation, and an Accuracy (Acc) of 82.3\% for lane line segmentation.

    \item TriLiteNet is deployed on embedded devices such as Jetson Xavier and Jetson TX2. It showcases real-time inference capabilities with low latency and optimal power consumption, making it suitable for practical applications.
\end{itemize}

The remainder of this paper is organized as follows: Section \ref{sec:related} provides a summary of related work. Subsequently, Section \ref{sec:method} introduces the Multi-Task Visual Perception model, TriLiteNet, which is capable of predicting three tasks: vehicle detection, drivable area segmentation, and lane line segmentation. Section \ref{sec:result} presents the experimental results of the TriLiteNet model on the BDD100K dataset. Finally, Section \ref{sec:conclusion} summarizes the key findings and discusses potential directions for future research.

\section{RELATED WORK}
\label{sec:related}

The development of perception models for autonomous vehicles has traditionally relied on single-task learning approaches, where individual tasks such as vehicle detection \cite{faster, yolov5, yolov8}, drivable area segmentation \cite{GCNet, pspnet, dltnet}, and lane line segmentation \cite{enet, enet-sad} are addressed independently. While these single-task models have achieved significant accuracy in their respective tasks, their reliance on separate models for each task leads to an exponential increase in computational requirements as the number of tasks grows. This limitation makes single-task approaches impractical for resource-constrained environments, such as embedded systems in autonomous vehicles.

To overcome these challenges, multi-task learning \cite{yolop, yolopv2, yolopv3, YOLOPX, ayolom, hybridnets,twin,twinplus} has emerged as a promising solution. Multi-task learning simultaneously optimizes multiple tasks, reducing redundancy and enhancing generalization by leveraging a shared encoder and task-specific decoders. This approach not only minimizes computational costs by avoiding the need for separate encoders but also enables the efficient sharing of features across tasks. For instance, features extracted for vehicle detection can also benefit segmentation tasks—a synergy that single-task models often fail to exploit. Consequently, multi-task learning is particularly well-suited for real-time autonomous driving systems, where computational efficiency and high performance are paramount.

Despite the impressive results of previous multi-task models in concurrently performing vehicle detection and segmentation tasks, many of these architectures suffer from significant limitations. Modern multi-task models often require tens of millions of parameters and substantial floating-point operations (FLOPs), making them unsuitable for environments with strict latency and energy constraints. Additionally, many current multi-task models utilize object detection backbones as the shared encoder, such as CSPDarknet in YOLOP \cite{yolov4scaled}, ELAN-Net in YOLOPv2, YOLOPX, and YOLOPv3 \cite{yolov7}, or CSPDarknet-c2f in A-YOLOM \cite{yolov8}. While these designs excel in vehicle detection, they are often suboptimal for segmentation tasks. Moreover, the performance of these models is typically evaluated on high-end hardware like the RTX 3090 or Tesla V100, which does not accurately reflect their feasibility for deployment in real-world autonomous driving systems. Previous works \cite{twin,twinplus} have addressed these issues by proposing lightweight multi-task models optimized for real-time inference on embedded devices. These models focus on drivable area and lane line segmentation tasks and demonstrate promising results in low-cost autonomous driving scenarios. However, they are limited to segmentation tasks and lack support for object detection, necessitating the integration of additional models to achieve comprehensive functionality.

To address these gaps, there is a pressing need for lightweight multi-task models that balance computational efficiency and performance while being deployable on resource-constrained systems. This study introduces TriLiteNet, a novel architecture designed to achieve this balance by efficiently handling multiple perception tasks, including vehicle detection, drivable area segmentation, and lane line segmentation. The proposed model is optimized for resource-constrained environments, such as embedded systems, and achieves real-time inference capabilities while maintaining high accuracy. Extensive evaluations of embedded devices demonstrate the practicality of TriLiteNet for real-world applications, particularly in autonomous driving scenarios with demanding deployment conditions.

\begin{figure}[!b]
    \includegraphics[width=0.47\textwidth]{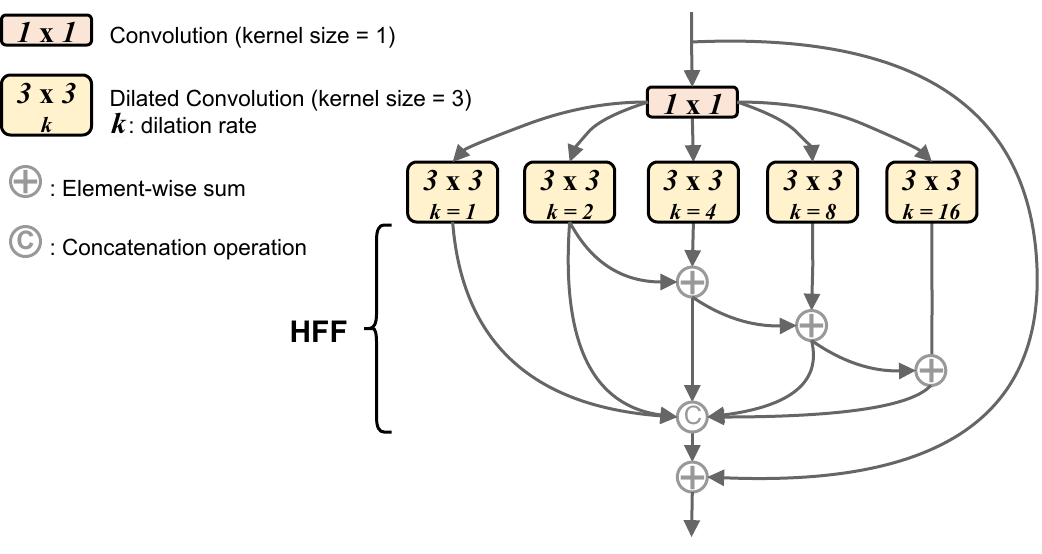}
    \caption{Design of the ESP module \cite{espnet}. The 1$\times$1 convolution reduces the dimensionality of the feature maps before feeding them into dilated convolutions with different dilation rates. The effective receptive field, introduced from the original ESP design, utilizes a hierarchical feature fusion (HFF) mechanism. A skip connection is incorporated between the input and output to improve information flow.}
    \label{fig:esp}
\end{figure}

\section{METHODOLOGY}
\label{sec:method}

\subsection{Network Architecture}

Based on the limitations of previous models, this study proposes an end-to-end model named TriLiteNet, which can perform multiple tasks at a low computational cost. TriLiteNet comprises a shared encoder and three task-specific decoders for Vehicle Detection, Drivable Area Segmentation, and Lane Line Segmentation.
As illustrated in Figure \ref{fig:pipeline}, an input image $I \in \mathcal{R}^{H \times W \times 3}$ is first processed by the encoder to extract multi-scale features $C_1$, $C_2$, $C_3$, $C_4$, and $C_5$ at resolutions of 1/2, 1/4, 1/8, 1/16, and 1/32 of the input resolution, respectively. Subsequently, features $C_3$, $C_4$, and $C_5$ are processed through Spatial Pyramid Pooling (SPP) and the proposed LitePAN, producing output features $P_3$, $P_4$, and $P_5$, which are then fed into the object detection head. The LitePAN module is inspired by the design of the Path Aggregation Network (PAN) with one top-down pathway and one bottom-up pathway, both simplified using Depth-wise Separable Convolutions. Additionally, the feature $C_3$ is processed through the Partial Class Activation Attention (PCAA) \cite{pcaa} module before being sent to the decoders for Drivable Area Segmentation and Lane Line Segmentation tasks. The two decoders for these segmentation tasks share a unified design that ensures low computational costs while maintaining performance. Finally, TriLiteNet integrates the outputs from its multi-task heads to provide a comprehensive panoramic driving perception result.

\subsubsection{Encoder}


The encoder of TriLiteNet is based on the ESPNet architecture \cite{espnet}, which utilizes the Efficient Spatial Pyramid (ESP) block as its foundation. The ESP block leverages parallel dilated convolutions to learn multi-scale representations effectively. The design of the ESP block, shown in Figure \ref{fig:esp}, consists of four main stages: Reduce, Split, Transform, and Merge. Specifically, the ESP block first reduces the dimensionality of the high-resolution input feature map using $1 \times 1$ convolutions. Subsequently, dilated convolutions with different dilation rates are applied in parallel, replacing standard convolution kernels. The feature map is then downsampled through the Stride ESP block, where 1$\times$1 convolutions are replaced with 3$\times$3 strided convolutions. In the ESPNet architecture, the ESP blocks are applied $N$ times, and the output of the current block is concatenated with the output of the previous block to preserve information. In this work, we propose replacing the dilated convolutions in the ESP block with depthwise dilated separable convolutions \cite{espnetv2} while retaining the Stride ESP block. This new combination is referred to as Depth-wise ESP. Thanks to the computational efficiency of depthwise separable convolutions and their proven compatibility with ESPNet in previous studies \cite{twinplus}, this modification significantly reduces the computational cost, specifically by $N$ times compared to the standard ESP blocks. As a result, Depth-wise ESP optimizes computational efficiency and maintains the ability to capture features at multiple scales through dilated convolutions, ensuring that the model achieves optimal representational learning performance.

\begin{figure*}
    \includegraphics[width=\textwidth]{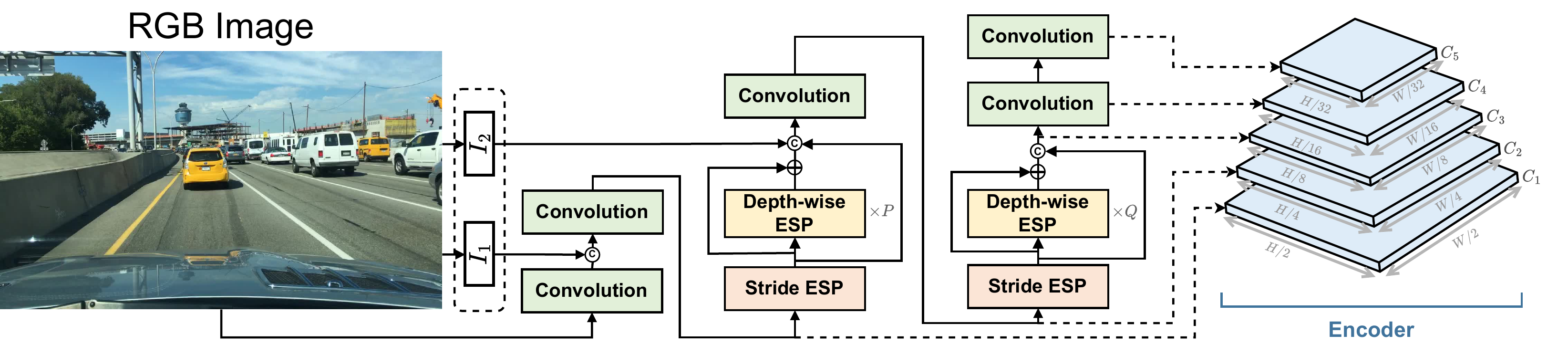}
    \caption{The encoder design of TriLiteNet comprises Stride ESP and Depth-wise ESP blocks, alongside standard convolution operations. During the encoding process, input images at different resolutions are integrated using skip connections. The output of the encoder includes feature maps at multiple scales, enabling the model to capture information at various levels of detail.}
    \label{fig:encoder}
\end{figure*}

The encoder in previous studies based on ESPNet\cite{twin, twinplus, espnet} transforms an input of size \( H \times W \) into a feature map with a minimum size of \( \frac{1}{8} \) of the input dimensions before proceeding to the decoding stage. In contrast, the TriLiteNet encoder is designed to generate feature maps at multiple scales, denoted as \( \{C_1, C_2, C_3, C_4, C_5\} \). These multi-scale feature maps enable the model to capture information at different levels of abstraction, enhancing its feature representation capability. Specifically, in object detection tasks, utilizing feature maps at scales such as \( \frac{1}{8} \), \( \frac{1}{16} \), and \( \frac{1}{32} \) of the original input size is essential, as it allows the model to detect and handle objects of varying sizes efficiently. To achieve this, in the TriLiteNet encoder, we add two strided convolution operations to produce features \( C_4 \in \mathcal{R}^{\frac{H}{16} \times \frac{W}{16}} \) and \( C_5 \in \mathcal{R}^{\frac{H}{32} \times \frac{W}{32}} \). The design of the encoder is illustrated in Figure~\ref{fig:encoder}, where: \( I_1 \in \mathcal{R}^{3 \times \frac{H}{2} \times \frac{W}{2}}, \quad I_2 \in \mathcal{R}^{3 \times \frac{H}{4} \times \frac{W}{4}} \). Here, \( I_1 \) and \( I_2 \) represent down-sampled versions of the input image, obtained by average pooling. These downsampled inputs provide additional contextual information from the original image, improving feature extraction during encoding. In the encoder, the Depth-wise ESP block is initially repeated \( P \) times, while subsequent Depth-wise ESP blocks are performed \( Q \) times. The hyperparameters \( P \) and \( Q \) are defined based on the model configuration, as detailed in Table~\ref{table:config}.

\subsubsection{Decoder for detection task}

\begin{figure}[!b]
    \includegraphics[width=0.5\textwidth]{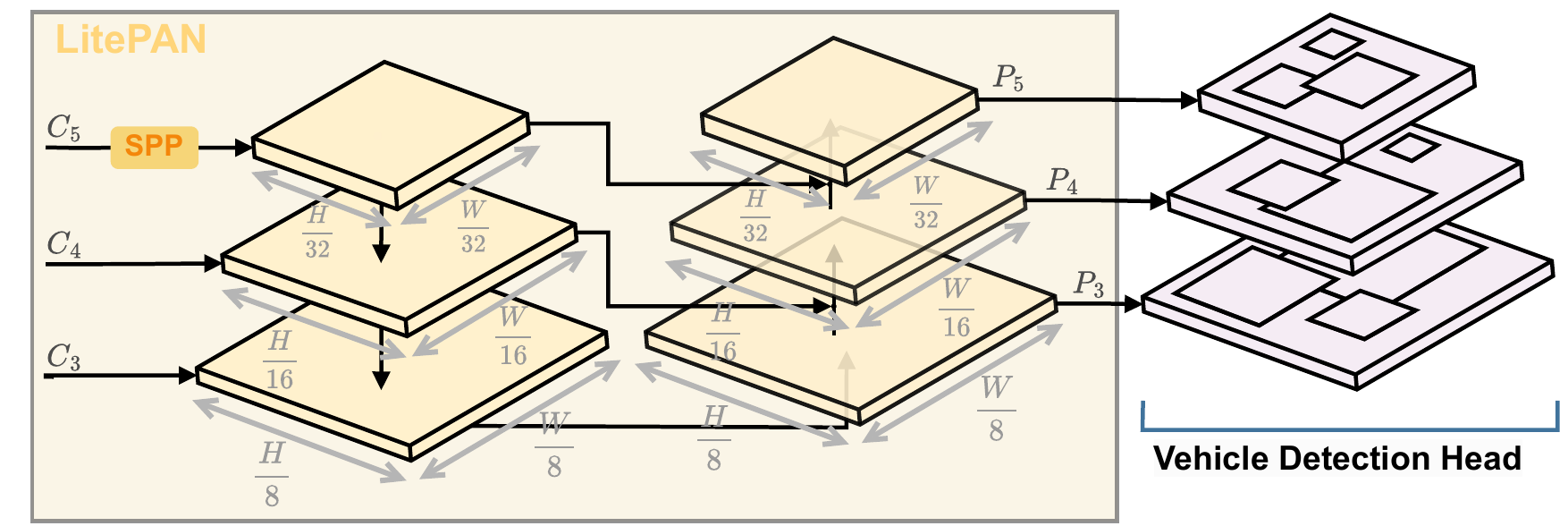}
    \caption{The output of the TriLitePAN is fed into the object detection head, where each grid cell of the multi-scale feature map is assigned three predefined anchors with different aspect ratios. The detection head then makes predictions based on these anchors.}
    \label{fig:det_decoder}
\end{figure}

\begin{figure}
    \includegraphics[width=0.5\textwidth]{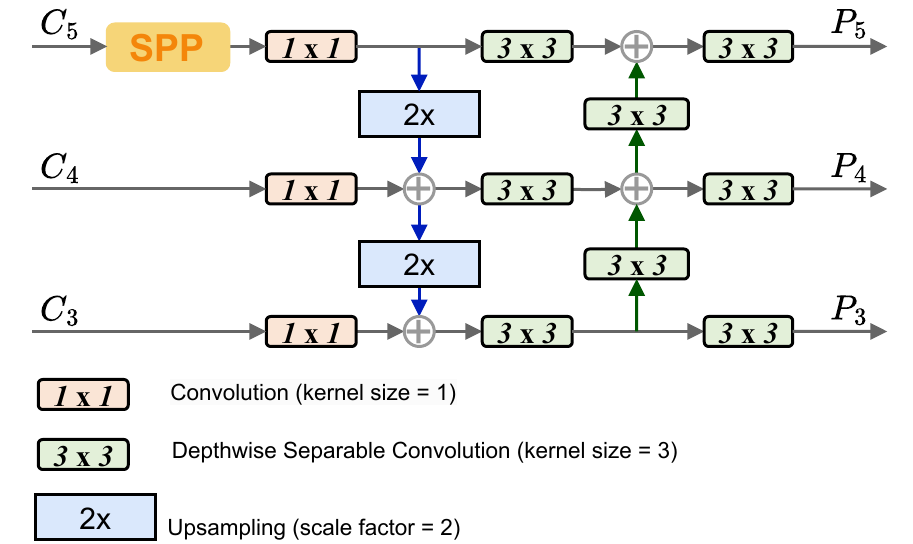}
    \caption{The proposed design of LitePAN is based on the architecture of PAN \cite{pan}, incorporating a top-down pathway and a bottom-up pathway. Our design centers on Depth-wise Separable Convolutions, which offer low computational cost while maintaining performance efficiency.}
    \label{fig:litepan}
\end{figure}

TriLiteNet leverages features at different resolutions in the decoder for object detection tasks.  The simple design of the detection head is illustrated in Figure \ref{fig:det_decoder}. Utilizing multi-resolution features enables the model to detect objects of varying sizes effectively, enhancing its versatility and accuracy in diverse scenarios. The multi-scale features are processed through LitePAN, starting with the feature map at a resolution of H/32 $\times$ W/32, which is refined using Spatial Pyramid Pooling (SPP) before being passed through LitePAN. The SPP module generates and merges features across multiple scales, while LitePAN integrates features at various semantic levels to produce multi-scale and semantically rich representations. LitePAN is optimized for computational efficiency and is based on the Path Aggregation Network (PAN). The design of LitePAN is illustrated in Figure \ref{fig:litepan}. Initially, LitePAN uses 1$\times$1 convolutions to adjust the channel dimensions of the feature maps. Then, feature transformations and extractions are performed using depthwise separable convolutions and upsampling, with feature maps of the same resolution combined through element-wise addition. Like PAN, our design includes bottom-up and top-down pathways, enhancing feature representation by effectively aggregating information across various levels and scales and improving spatial details and semantic understanding. LitePAN is the neck in object detection tasks, acting as a bridge between the encoder and the detection head. Through successive layers, it enhances and combines features from the backbone, providing enriched representations for the head to achieve better detection performance. 

To effectively handle small and overlapping objects in the BDD100K \cite{bdd100k} dataset, we adopt an anchor-based approach instead of an anchor-free one. To enhance the adaptability of the anchor-based approach, we integrate an auto anchor mechanism \cite{yolov4} based on K-means. This mechanism dynamically determines the optimal anchor box sizes, reducing manual tuning while maintaining efficiency. Each grid cell of the multi-scale feature map produced by the LitePAN is assigned three anchors with varying aspect ratios. The detection head then predicts the positional offsets, width and height ratios, the probability of each class, and the confidence of the predictions.

\subsubsection{Decoder for segmentation tasks}

To generate accurate segmentation map outputs for two tasks, lane line segmentation, and drivable area segmentation, we propose an upsampling design that transforms the feature map \( C_3 \in \mathcal{R}^{\frac{H}{8} \times \frac{W}{8}} \) into two segmentation maps, \( O_{da} \) and \( O_{ll} \), both having the same size as the input image \( H \times W \). Here, \( O_{da} \) corresponds to the drivable area segmentation, while \( O_{ll} \) corresponds to the lane line segmentation. Specifically, this transformation is not performed through dimensional splitting but rather through two independent segmentation heads with separate weights.

\begin{figure}[!b]
    \centering
    \begin{subfigure}[b]{0.5\textwidth}
        \centering
        \includegraphics[width=\textwidth]{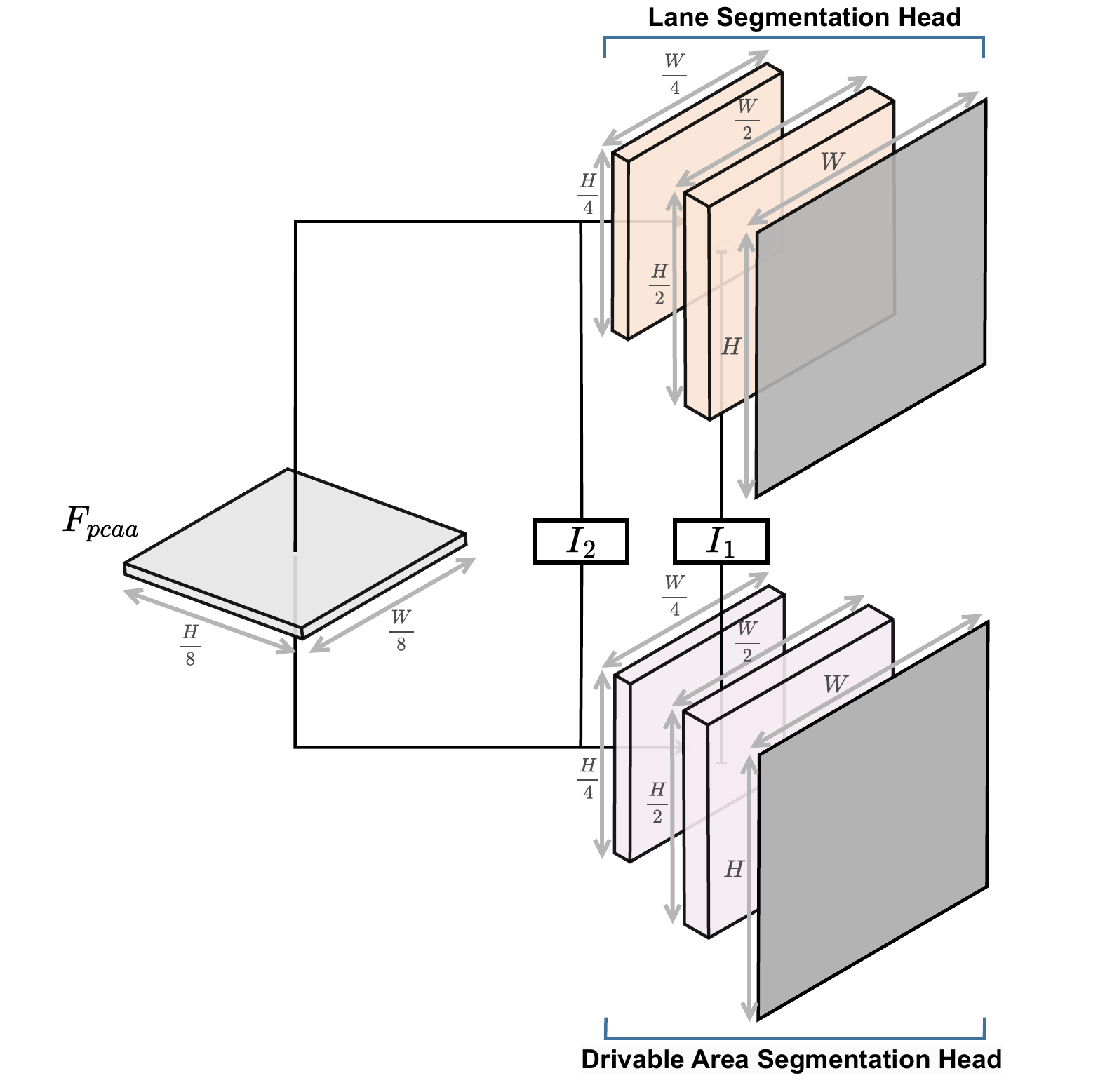}
        \caption{Overview architecture of segmentation decoders: These two heads share a similar design, receiving input from the PCAA block. During upsampling, the lower-resolution input is concatenated with feature maps of corresponding resolutions through skip connections, enhancing information flow throughout the decoding process.}
        \label{fig:seghead_a}
    \end{subfigure}
    
    \begin{subfigure}[b]{0.5\textwidth}
        \centering
        \includegraphics[width=\textwidth]{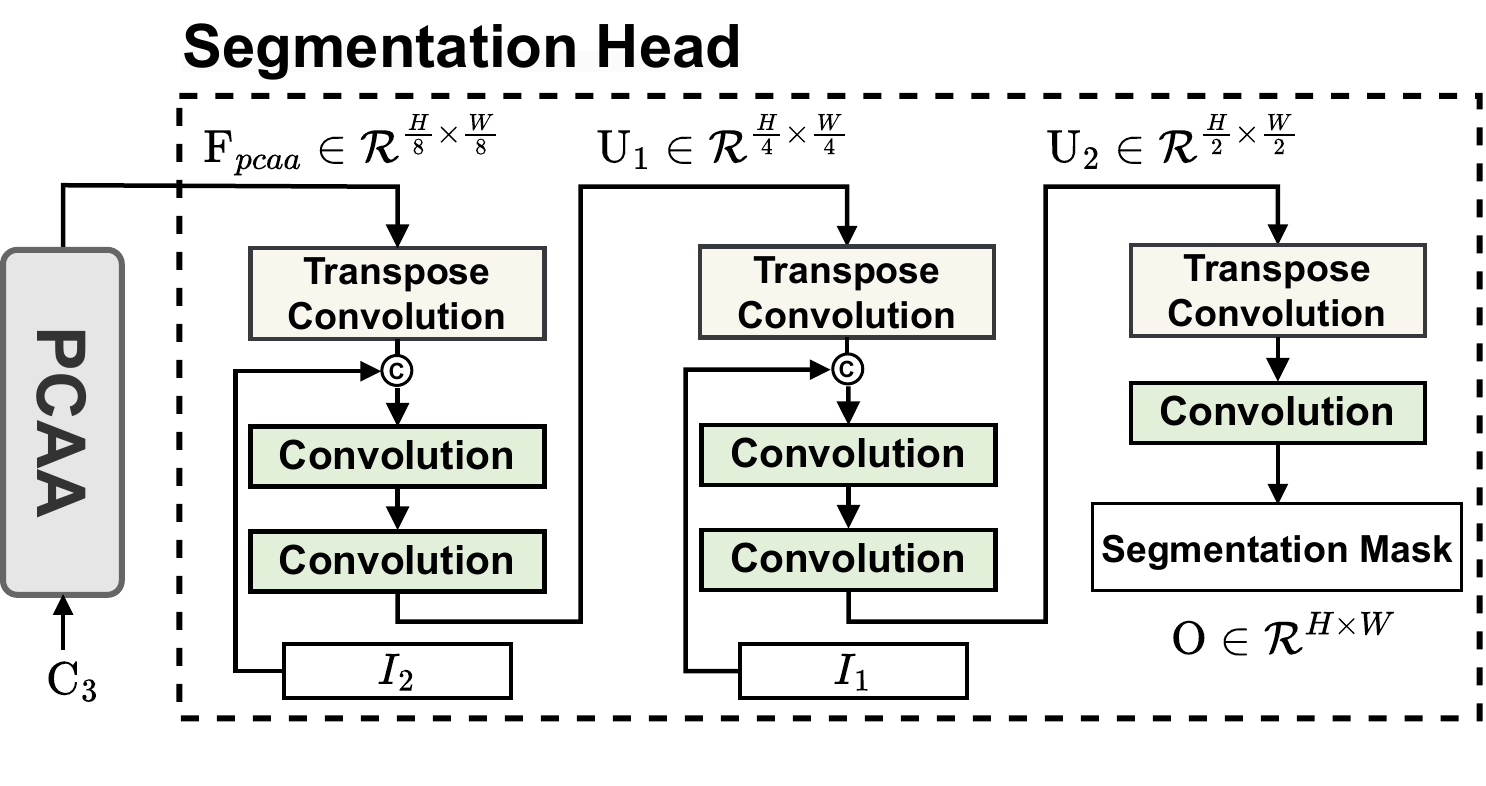}
        \caption{The segmentation head's detailed design: It takes the output of the PCAA module, and the segmentation head for each task progressively upsamples the feature map at each stage using Transposed Convolutions and Convolutions to increase its resolution.}
        \label{fig:seghead_b}
    \end{subfigure}
    
    \caption{Design of segmentation decoders}
    \label{fig:two_images_vertical}
\end{figure}

First, the feature map \( C_3 \), extracted from the encoder at a resolution of \( \frac{H}{8} \times \frac{W}{8} \), is passed through the Partial Class Activation Attention (PCAA) module to generate the feature map \( F_{pcaa} \in \frac{H}{8} \times \frac{W}{8} \). The PCAA module gathers local and global class-level representations for attention calculation, thereby improving segmentation accuracy and efficiency by focusing on critical regions, such as the drivable area and lane markings. Next, $F_{pcaa}$ is fed into two identical but separately weighted decoder heads: the Lane Line Segmentation Head and the Drivable Area Segmentation Head. Each head applies a series of transposed convolutions for upsampling, followed by convolutions for feature refinement. The decoder architecture for the two segmentation tasks of TriLiteNet is depicted in Figure~\ref{fig:seghead_a}. During the upsampling process, skip connections are utilized to integrate the downsampled inputs \( I_1 \in \mathcal{R}^{3 \times \frac{H}{2} \times \frac{W}{2}} \) and \( I_2 \in \mathcal{R}^{3 \times \frac{H}{4} \times \frac{W}{4}} \). These connections help preserve feature consistency with the input image while enhancing prediction performance. Both segmentation heads follow the same architecture but employ distinct sets of weights, enabling the model to learn task-specific patterns without interference. The combination of PCAA and a simple decoder architecture has previously demonstrated strong performance in autonomous vehicle segmentation tasks \cite{twinplus}. Finally, the two segmentation heads perform progressive upsampling through multiple steps to generate the outputs \( O_{da}, O_{ll} \in  \mathcal{R}^{2 \times H \times W} \), corresponding to the predictions for each task. The design of the segmentation heads is illustrated in Figure~\ref{fig:seghead_b}, where the input is \( F_{pcaa} \) and the outputs are \( O_{da} \) and \( O_{ll} \).

    

\begin{table}[]
\centering
\caption{TriLiteNet Configurations: \textbf{Aspect Ratio} refers to the ratio of the feature map's resolution relative to the input image, while \textbf{Output Channels} denote the number of output channels in the feature maps corresponding to each configuration.}
\label{table:config}
\begin{tabular}{lllccc}
\toprule
\multicolumn{2}{l}{\multirow{2}{*}{\textbf{Feature maps}}} & \multirow{2}{*}{\textbf{Aspect Ratio}} & \multicolumn{3}{c}{\textbf{Ouput Chanels}}                        \\ \cmidrule{4-6} 
\multicolumn{2}{l}{}                                       &                                      & \textbf{Tiny}       & \textbf{Small}       & \textbf{Base}        \\ \midrule
\multirow{5}{*}{\rotatebox[origin=c]{90}{\begin{tabular}[c]{@{}c@{}}\textbf{Encoder}\end{tabular}}}          & $C_1$             & 1/2     & 8                   & 16                   & 32                   \\
                                          & $C_2$             & 1/4             & 32                  & 64                   & 128                  \\
                                          & $C_3$             & 1/8              & 64 & 128 & 256 \\
                                          & $C_4$             & 1/16           &   64 & 128 & 256 \\
                                          & $C_5$             & 1/32           &  64 & 128 & 256 \\ \midrule
\addlinespace[0.3em]
\multirow{3}{*}{\rotatebox[origin=c]{90}{\begin{tabular}[c]{@{}c@{}}\textbf{Detection}\\\textbf{Decoder}\end{tabular}}}        & $P_3$             & 1/8              & 32 & 64  & 128 \\
\addlinespace[0.3em]
                                          & $P_4$             & 1/16           & 32 & 64  & 128 \\
\addlinespace[0.3em]
                                          & $P_5$             & 1/32        & 32 & 64  & 128 \\
\addlinespace[0.3em]
\midrule \addlinespace[0.35em]
\multirow{4}{*}{\rotatebox[origin=c]{90}{\begin{tabular}[c]{@{}c@{}}\textbf{Segmentation}\\\textbf{Decoder}\end{tabular}}}     & $F_{pcca}$     & 1/8              & 16                  & 32                   & 64                   \\ \addlinespace[0.35em]
                                          & $U_{1(da,ll)}$   & 1/4             & 8                   & 16                   & 32                   \\ \addlinespace[0.35em]
                                          & $U_{2(da,ll)}$   & 1/2             & 4                   & 8                    & 16                   \\ \addlinespace[0.35em]
                                          & $O_{da,ll}$    &     1                         & 2                   & 2                    & 2                    \\ \addlinespace[0.35em] \midrule
\multicolumn{3}{l}{\textbf{P}, \textbf{Q}}                                                        & 1, 1                & 2, 3                 & 3, 5                 \\ \bottomrule
\end{tabular}%
\end{table}

\subsection{Model Configurations}

In this paper, we propose TriLiteNet with three configurations: TriLiteNet$_{tiny}$, TriLiteNet$_{small}$, and TriLiteNet$_{base}$. All three configurations share the same overall architecture, with the primary differences being the number of kernels in the convolutional layers and the two hyperparameters \( P \) and \( Q \) (the number of iterations of the Depth-wise ESP blocks). Table~\ref{table:config} provides a detailed breakdown of the output channel sizes and resolutions for the feature maps in the Encoder (\( C_1, C_2, C_3, C_4, C_5 \)), Detection Decoder (\( P_3, P_4, P_5 \)), and Segmentation Decoder (\( F_{pcaa}, U_1, U_2, O \)) for both segmentation tasks. The model complexity increases progressively in the order of \textit{tiny}, \textit{small}, and \textit{base}, enabling a balance between computational cost and task performance.

\subsection{Loss function}

Since there are three independent outputs for different tasks, our composite loss function consists of three components. The object detection loss calculates the weighted sum of the classification loss $\mathcal{L}_{det}^{cls}$, object loss $\mathcal{L}_{det}^{obj}$, and regression loss $\mathcal{L}_{det}^{reg}$:

\begin{equation}
    \mathcal{L}_{det} = \omega_{cls} \mathcal{L}_{det}^{cls} + \omega_{obj} \mathcal{L}_{det}^{obj} + \omega_{reg} \mathcal{L}_{det}^{reg}
\end{equation}

where $\mathcal{L}_{det}^{cls}$ and $\mathcal{L}_{det}^{obj}$ are Binary Cross-Entropy (BCE) losses, with $\mathcal{L}_{det}^{cls}$ used for class classification and $\mathcal{L}_{det}^{obj}$ for object confidence. The regression loss $\mathcal{L}_{det}^{reg}$, based on IoU loss, measures the distance of overlap rate, aspect ratio, and scale similarity between the predicted and actual bounding boxes. The hyperparameters $\omega_{cls}$, $\omega_{obj}$, and $\omega_{reg}$ are set to 0.5, 1.0, and 0.05, respectively, to balance the terms in $\mathcal{L}_{det}$, as tuned in YOLOP \cite{yolop}.

To train the drivable area segmentation and lane line segmentation tasks, we combine Focal Loss \cite{focal} and Tversky Loss \cite{tversky} by summing them as follows:

\begin{equation}
    \mathcal{L}_{da} = \mathcal{L}_{da}^{focal} + \mathcal{L}_{da}^{tversky}
\end{equation}

\begin{equation}
    \mathcal{L}_{ll} = \mathcal{L}_{ll}^{focal} + \mathcal{L}_{ll}^{tversky}
\end{equation}

where $\mathcal{L}^{focal}$ increases the weight for hard-to-classify pixels and reduces the impact of easily classified ones, effectively addressing the class imbalance. Meanwhile, $\mathcal{L}^{tversky}$ handles class imbalance in segmentation tasks by controlling the influence of false positives and false negatives. The hyperparameters used for these loss functions are finely tuned as follows: $\alpha$ = 0.25, $\gamma$ = 2.0 for $\mathcal{L}^{focal}$ across both tasks, $\alpha$ = 0.7, $\beta$ = 0.3 for $\mathcal{L}_{da}^{tversky}$, and $\alpha$ = 0.9, $\beta$ = 0.1 for $\mathcal{L}_{ll}^{tversky}$.

The final loss function is calculated through a weighted sum as follows:

\begin{equation}
    \mathcal{L}_{final} = \omega_{del} \mathcal{L}_{det} + \omega_{da} \mathcal{L}_{da} + \omega_{ll} \mathcal{L}_{ll}
\end{equation}

where $\mathcal{L}_{det}$, $\mathcal{L}_{da}$, and $\mathcal{L}_{ll}$ represent the loss functions for object detection, drivable area segmentation, and lane line segmentation tasks, respectively. The weights are fine-tuned with $\mathcal{L}_{det}$ = 1, $\mathcal{L}_{da}$ = 0.3, and $\mathcal{L}_{ll}$ = 0.3.

\section{Experimental Results}
\label{sec:result}

\begin{algorithm}
\caption{The training phase of TriLiteNet.}
\label{alg_train}
\begin{algorithmic}[1] 
\REQUIRE Target end-to-end network $F$ with parameters $\Theta$; Training dataset $\tau_{train}$; Validation dataset $\tau_{val}$
\ENSURE Well-trained network: $F(x; \Theta)$
\STATE Initialize the parameters $\Theta$
\FOR {epoch: 1 $\rightarrow$ epochs}
    
    \FOR{each batch $(x_{train}, y_{train})$ in $\tau_{train}$}

            \STATE $y_\text{det}, y_\text{da}, y_\text{ll} \gets y_{train}$
            \vspace{0.4em}
            \STATE $\hat{y}_\text{det}, \hat{y}_\text{da}, \hat{y}_\text{ll} \gets F(x_{train})$
            \vspace{0.4em}
            \STATE $\mathcal{L}_\text{det} \gets \mathcal{L}_\text{det}^{cls}(y_\text{det}, \hat{y}_\text{det})$ + $\mathcal{L}_\text{det}^{obj}(y_\text{det}, \hat{y}_\text{det})$ 
            \STATE \quad \quad \quad +  $\mathcal{L}_\text{det}^{reg}(y_\text{det}, \hat{y}_\text{det})$
            \vspace{0.4em}
            \STATE $\mathcal{L}_\text{da} \gets \mathcal{L}_\text{da}^{focal}(y_\text{da}, \hat{y}_\text{da})$ + $\mathcal{L}_\text{da}^{tversky}(y_\text{da}, \hat{y}_\text{da})$
            \vspace{0.4em}
            \STATE $\mathcal{L}_\text{ll} \gets \mathcal{L}_\text{ll}^{focal}(y_\text{ll}, \hat{y}_\text{ll})$ + $\mathcal{L}_\text{ll}^{tversky}(y_\text{ll}, \hat{y}_\text{ll})$
            \vspace{0.4em}
            \STATE $\mathcal{L} \gets \mathcal{L}_\text{det} + \mathcal{L}_\text{da} + \mathcal{L}_\text{ll}$
            \vspace{0.4em}
            \STATE $\Theta \gets {argmin}_{\Theta}\mathcal{L}$
            \vspace{0.4em}
            \STATE $\Theta \gets {\text{Update}}_{\text{EMA}}\Theta$
        \ENDFOR

    \ENDFOR
\STATE  \RETURN $F(x; \Theta)$
\end{algorithmic}
\end{algorithm}

\subsection{Training settings}

\subsubsection{Dataset}

We train and evaluate the model on the BDD100K dataset, which includes training, validation, and test sets with 70k, 10k, and 20k images, respectively. The dataset comprises various images from different weather conditions and scenes, making it a robust and generalizable dataset for autonomous driving tasks. The model is trained on the training set and evaluated on the validation set, as the test set is not publicly available. The original images are resized from 1280 $\times$ 720 pixels to 640 $\times$ 384 pixels. Four vehicle classes (car, truck, bus, and train) are combined into a single ``vehicle'' class for the object detection task. Additionally, we combine the direct drivable area and the alternative drivable area into a single drivable region for the drivable area segmentation task. Furthermore, we adjust the lane line width to 8 pixels in the training set while maintaining a width of 2 pixels in the validation set. These standard practices are adopted by prior works \cite{enet-sad,yolop,yolopv2,yolopv3,YOLOPX,ayolom} to ensure fairness in comparisons.

\subsubsection{Implement Details}

We implement the TriLiteNet model using the PyTorch framework \cite{pytorch}. Our proposed model is trained with the AdamW optimizer \cite{adamw} over 200 epochs with batch sizes of 16 images. The initial learning rate, $\beta_1$, and $\beta_2$ are set to 0.001, 0.937, and 0.999, respectively, for optimization. Learning rate scheduling with warm-up and cosine annealing is applied to converge faster and better. Importantly, no pre-trained models are used for fine-tuning. During the training process, we fully utilize the EMA (Exponential Moving Average) \cite{ema} model as the final inference model. The step-by-step procedure of the training method is presented in Algorithm \ref{alg_train}. All experiments are conducted on the device with an RTX 4090 GPU and an Intel(R) Core(TM) i9-10900X processor.

\subsubsection{Evaluation Metrics}

In line with previous studies \cite{yolop, yolopv2, YOLOPX, yolopv3, ayolom}, we adopt Recall and mAP@0.5 as evaluation metrics for object detection tasks. For segmentation tasks, we use mean Intersection over Union (mIoU) to assess drivable area segmentation and Pixel Accuracy (Acc) and Intersection over Union (IoU) for lane line segmentation. Specifically, instead of conventional pixel accuracy, we utilize balanced accuracy \cite{ayolom}, which provides a fairer evaluation by accounting for the accuracy of each class. Additionally, we evaluate the model based on its number of parameters and Floating-point operations (FLOPs). Consistent with definitions in prior works \cite{espnet, resnet}, FLOPs refer to the number of multiply-add operations. To ensure a fair comparison, the FLOPs are recalculated uniformly for all models under evaluation.

\begin{table}[]
\caption{The comparison results of the computational cost.}
\label{tab:cost}
\begin{tabular}{lccccc}

\toprule
\multirow{2}{*}{\textbf{Model}} & \multicolumn{3}{c}{\textbf{FPS (with Batch Size)}} & \multirow{2}{*}{\textbf{Params}} & \multirow{2}{*}{\textbf{FLOPs}} \\ \cmidrule{2-4}
                                & 1            & 8         & 32            &                                      &                                \\ \midrule
YOLOP\cite{yolop}                           & 63           & 470                 & 732           & 7.9M                                  & 9.38G                           \\
YOLOPv2\cite{yolopv2}                         & 88           & 261                 & 324           & 38.9M                                 & \xmark                              \\
YOLOPv3\cite{yolopv3}                         & 46           & 243                 & 268           & 30.2M                                 & 35.38G                          \\
Hybridnets\cite{hybridnets}                      & 8            & 53                  & 121           & 13.8M                                 & 25.69G                          \\
YOLOPX\cite{YOLOPX}                          & 49           & 199                 & 262           & 32.9M                                 & 44.86G                          \\
AYOLOM (n)\cite{ayolom}                      & 117          & 552                 & 620           & 13.61M                                & 6.66G                           \\
AYOLOM (s)\cite{ayolom}                      & 112          & 381                 & 348           & 4.43M                                 & 19.22G                          \\ \midrule

TriLiteNet$_{tiny}$               & 185          & 1340                & 3397          & 0.15M                                 & 0.55G                           \\ 
TriLiteNet$_{small}$                      & 151          & 1081                & 1641          & 0.59M                                 & 1.99G                           \\
TriLiteNet$_{base}$                      & 105          & 727                & 760          & 2.35M    & 7.72G                           \\

\bottomrule
\end{tabular}%
\end{table}

\begin{table*}[]

\centering
\caption{Performance Comparison with TriLiteNet and other models. \xmark \text{ } denotes tasks for which the model has not been trained, while ``--'' indicates results that are not available. To compare other models with TriLiteNet, we use the symbols \pmb{$\uparrow$} and \pmb{$\downarrow$} to indicate the difference in performance metrics. For Params and FLOPs, we calculate how many times other models exceed TriLiteNet, denoted by the symbol ({\color{red}\pmb{$\times$}} or {\color{blue}\pmb{$\times$}}). \textcolor{red}{\textbf{Red}} values indicate that the model outperforms TriLiteNet, while \textcolor{blue}{\textbf{blue}} values indicate the opposite.}
\label{tab:comparison}
\begin{tabular}{llccccccc}
\toprule
 & \multirow{2}{*}{\textbf{Model}}                          & \multicolumn{2}{c}{\textbf{Vehicle Detection}}                          & \textbf{Drivable Area Segmentation}     & \multicolumn{2}{c}{\textbf{Lane Line Segmentation}}                          & \multirow{2}{*}{\textbf{Params}} & \multirow{2}{*}{\textbf{FLOPs}} \\ \cmidrule(lr){3-4}\cmidrule(lr){5-5}\cmidrule(lr){6-7}
&                                                          & \textbf{Recall (\%)}               & \textbf{mAP50 (\%)}                & \textbf{mIoU (\%)}                 & \textbf{Acc (\%)}                 & \textbf{IoU (\%)}                   &                                      &                                     \\ \cmidrule{2-9}
& TriLiteNet$_{base}$ (Ours)                                        & 85.6                               & 72.3                               & 92.4                               & 82.3                              & 29.8                                & 2.35M                                 & 7.72G                                \\ \midrule
\multirow{12}{*}{\rotatebox[origin=c]{90}{\begin{tabular}[c]{@{}c@{}}\textbf{Object Detection} \end{tabular}}} & \multirow{2}{*}{YOLOv5s \cite{yolov5}}                   & 86.8                               & 77.2                               & \multirow{2}{*}{\xmark}            & \multicolumn{2}{c}{\multirow{2}{*}{\xmark}}                             & –                                    & –                                   \\
&                                                          & \textcolor{blue}{$\uparrow$ 1.2}   & \textcolor{blue}{$\uparrow$ 4.9}   &                                    & \multicolumn{2}{c}{}                                                    &                                      &                                     \\ \cmidrule{2-9}
& \multirow{2}{*}{Faster RCNN \cite{faster}}               & 77.2                               & 55.6                               & \multirow{2}{*}{\xmark}            & \multicolumn{2}{c}{\multirow{2}{*}{\xmark}}                             & –                                    & –                                   \\
&                                                          & \textcolor{red}{$\downarrow$ 8.4}  & \textcolor{red}{$\downarrow$ 16.7} &                                    & \multicolumn{2}{c}{}                                                    &                                      &                                     \\ \cmidrule{2-9}
& \multirow{2}{*}{MultiNet \cite{multinet}}                & 81.3                               & 60.2                               & \multirow{2}{*}{\xmark}            & \multicolumn{2}{c}{\multirow{2}{*}{\xmark}}                             & –                                    & –                                   \\
&                                                          & \textcolor{red}{$\downarrow$ 4.3}  & \textcolor{red}{$\downarrow$ 12.1} &                                    & \multicolumn{2}{c}{}                                                    &                                      &                                     \\ \cmidrule{2-9}
& \multirow{2}{*}{R-CNNP (DET) \cite{yolop}}               & 79.0                               & 67.3                               & \multirow{2}{*}{\xmark}            & \multicolumn{2}{c}{\multirow{2}{*}{\xmark}}                             & –                                    & –                                   \\
&                                                          & \textcolor{red}{$\downarrow$ 6.6}  & \textcolor{red}{$\downarrow$ 5.0}  &                                    & \multicolumn{2}{c}{}                                                    &                                      &                                     \\ \cmidrule{2-9}
& \multirow{2}{*}{YOLOP (DET) \cite{yolop}}                & 88.2                               & 76.9                               & \multirow{2}{*}{\xmark}            & \multicolumn{2}{c}{\multirow{2}{*}{\xmark}}                             & –                                    & –                                   \\
&                                                          & \textcolor{blue}{$\uparrow$ 2.6}   & \textcolor{blue}{$\uparrow$ 4.6}   &                                    & \multicolumn{2}{c}{}                                                    &                                      &                                     \\ \cmidrule{2-9}
& \multirow{2}{*}{YOLOv8n (DET) \cite{yolov8}}             & 82.2                               & 75.1                               & \multirow{2}{*}{\xmark}            & \multicolumn{2}{c}{\multirow{2}{*}{\xmark}}                             & 3.16M                                 & –                                   \\ 
&                                                          & \textcolor{red}{$\downarrow$ 3.4}  & \textcolor{blue}{$\uparrow$ 2.8}   &                                    & \multicolumn{2}{c}{}                                                    & \textcolor{red}{$\times$ 1.34}       &                                     \\ \midrule
                                                         
\multirow{10}{*}{\rotatebox[origin=c]{90}{\begin{tabular}[c]{@{}c@{}}\textbf{Drivable Are Segmentation} \end{tabular}}} & \multirow{2}{*}{GCNet \cite{GCNet}}                      & \multicolumn{2}{c}{\multirow{2}{*}{\xmark}}                             & 82.07                              & \multicolumn{2}{c}{\multirow{2}{*}{\xmark}}                             & –                                    & –                                   \\
&                                                          &                                    &                                    & \textcolor{red}{$\downarrow$ 10.33} & \multicolumn{2}{c}{}                                                    &                                      &                                     \\  \cmidrule{2-9}
& \multirow{2}{*}{DNLNet \cite{dnlnet}}                    & \multicolumn{2}{c}{\multirow{2}{*}{\xmark}}                             & 84.35                              & \multicolumn{2}{c}{\multirow{2}{*}{\xmark}}                             & –                                    & –                                   \\
&                                                          &                                    &                                    & \textcolor{red}{$\downarrow$ 8.05} & \multicolumn{2}{c}{}                                                    &                                      &                                     \\  \cmidrule{2-9}
& \multirow{2}{*}{PSPNet \cite{pspnet}}                    & \multicolumn{2}{c}{\multirow{2}{*}{\xmark}}                             & 89.6                               & \multicolumn{2}{c}{\multirow{2}{*}{\xmark}}                             & –                                    & –                                   \\
&                                                          &                                    &                                    & \textcolor{red}{$\downarrow$ 2.8}  & \multicolumn{2}{c}{}                                                    &                                      &                                     \\  \cmidrule{2-9}
& \multirow{2}{*}{YOLOP (DA) \cite{yolop}}                 & \multicolumn{2}{c}{\multirow{2}{*}{\xmark}}                             & 92.0                               & \multicolumn{2}{c}{\multirow{2}{*}{\xmark}}                             & –                                    & –                                   \\
&                                                          &                                    &                                    & \textcolor{red}{$\downarrow$ 0.4}  & \multicolumn{2}{c}{}                                                    &                                      &                                     \\  \cmidrule{2-9}
& \multirow{2}{*}{YOLOv8n (DA) \cite{yolov8}}              & \multicolumn{2}{c}{\multirow{2}{*}{\xmark}}                             & 78.1                               & \multicolumn{2}{c}{\multirow{2}{*}{\xmark}}                             & 3.26M                                 & –                                   \\
&                                                          &                                    &                                    & \textcolor{red}{$\downarrow$ 14.3} & \multicolumn{2}{c}{}                                                    & \textcolor{red}{$\times$ 1.39}       &                                     \\ \midrule
\multirow{10}{*}{\rotatebox[origin=c]{90}{\begin{tabular}[c]{@{}c@{}}\textbf{Lane Line Segmentation} \end{tabular}}} & \multirow{2}{*}{Enet \cite{enet}}                        & \multicolumn{2}{c}{\multirow{2}{*}{\xmark}}                             & \multirow{2}{*}{\xmark}            & –                                 & 14.64                              & –                                    & –                                   \\
                                                         &                                    &                                    &                                    &                                   &  & \textcolor{red}{$\downarrow$ 15.16}                                      &                                     \\  \cmidrule{2-9}
& \multirow{2}{*}{ENet-SAD \cite{enet-sad}}                & \multicolumn{2}{c}{\multirow{2}{*}{\xmark}}                             & \multirow{2}{*}{\xmark}            & –                                 & 16.02                              & –                                    & –                                   \\
                                                         &                                    &                                    &                                    &                                   & & \textcolor{red}{$\downarrow$ 13.78}                                      &                                     \\  \cmidrule{2-9}
& \multirow{2}{*}{SCNN \cite{scnn}}                        & \multicolumn{2}{c}{\multirow{2}{*}{\xmark}}                             & \multirow{2}{*}{\xmark}            & –                                 & 15.84                              & –                                    & –                                   \\
                                                         &                                    &                                    &                                    &                                   & & \textcolor{red}{$\downarrow$ 13.96}                                      &                                     \\  \cmidrule{2-9}
& \multirow{2}{*}{YOLOP (LL) \cite{yolop}}                 & \multicolumn{2}{c}{\multirow{2}{*}{\xmark}}                             & \multirow{2}{*}{\xmark}            & –                                 & 26.5                               & –                                    & –                                   \\
                                                         &                                    &                                    &                                    &                                   &  & \textcolor{red}{$\downarrow$ 3.3}                                      &                                     \\  \cmidrule{2-9}
& \multirow{2}{*}{YOLOv8n (LL) \cite{yolov8}}              & \multicolumn{2}{c}{\multirow{2}{*}{\xmark}}                             & \multirow{2}{*}{\xmark}            & 80.5                              & 22.9                               & 3.26M                                 & –                                \\
&                                                         &                                    &                                    &                                    & \textcolor{red}{$\downarrow$ 1.8}  & \textcolor{red}{$\downarrow$ 6.9}   & \textcolor{red}{$\times$ 1.39}       &       \\ \midrule
\multirow{14}{*}{\rotatebox[origin=c]{90}{\begin{tabular}[c]{@{}c@{}}\textbf{Multi-tasking}\end{tabular}}} & \multirow{2}{*}{YOLOP \cite{yolop}}                      & 89.2                               & 76.5                               & 91.5                               & 84.8                              & 26.2                               & 7.9M                                  & 9.38G                                \\
&                                                          & \textcolor{blue}{$\uparrow$ 3.6}   & \textcolor{blue}{$\uparrow$ 4.2}   & \textcolor{red}{$\downarrow$ 0.9}  & \textcolor{blue}{$\uparrow$ 2.5}  & \textcolor{red}{$\downarrow$ 3.6}   & \textcolor{red}{$\times$ 3.36}       & \textcolor{red}{$\times$ 1.21}      \\  \cmidrule{2-9}
& \multirow{2}{*}{HybridNets \cite{hybridnets}}            & 92.8                               & 77.3                               & 90.5                               & –                                 & 31.6                               & 12.83M                                & 25.69G                               \\
&                                                          & \textcolor{blue}{$\uparrow$ 7.2}   & \textcolor{blue}{$\uparrow$ 5.0}   & \textcolor{red}{$\downarrow$ 1.9}  &                                   & \textcolor{blue}{$\uparrow$ 1.8}   & \textcolor{red}{$\times$ 5.46}       & \textcolor{red}{$\times$ 3.33}      \\  \cmidrule{2-9}
& \multirow{2}{*}{YOLOPv2 \cite{yolopv2}}                  & 91.1                               & 83.4                               & 93.2                               & –                                 & 27.25                              & 38.9M                                 & –                                   \\
&                                                          & \textcolor{blue}{$\uparrow$ 5.5}   & \textcolor{blue}{$\uparrow$ 11.1}  & \textcolor{blue}{$\uparrow$ 0.8}   &                                   & \textcolor{red}{$\downarrow$ 2.55}  & \textcolor{red}{$\times$ 16.55}      &                                     \\  \cmidrule{2-9}
& \multirow{2}{*}{A-YOLOM (n) \cite{ayolom}}               & 85.3                               & 78                                 & 90.5                               & 81.3                              & 28.2                               & 4.43M                                 & 6.66G                                \\
&                                                          & \textcolor{red}{$\downarrow$ 0.3}  & \textcolor{blue}{$\uparrow$ 5.7}   & \textcolor{red}{$\downarrow$ 1.9}  & \textcolor{red}{$\downarrow$ 1.0} & \textcolor{red}{$\downarrow$ 1.6}   & \textcolor{red}{$\times$ 1.89}       & \textcolor{blue}{$\times$ 0.86}      \\  \cmidrule{2-9}
& \multirow{2}{*}{A-YOLOM (s) \cite{ayolom}}               & 86.9                               & 81.1                               & 91                                 & 84.9                              & 28.8                               & 13.61M                                & 19.22G                               \\
&                                                          & \textcolor{blue}{$\uparrow$ 1.3}   & \textcolor{blue}{$\uparrow$ 8.8}   & \textcolor{red}{$\downarrow$ 1.4}  & \textcolor{blue}{$\uparrow$ 2.6}  & \textcolor{red}{$\downarrow$ 1.0}   & \textcolor{red}{$\times$ 5.79}       & \textcolor{red}{$\times$ 2.49}      \\  \cmidrule{2-9}
& \multirow{2}{*}{YOLOPX \cite{YOLOPX}}                    & 93.7                               & 83.3                               & 93.2                               & –                                 & 27.2                               & 32.9M                                 & 44.86G                               \\
 &                                                         & \textcolor{blue}{$\uparrow$ 8.1}   & \textcolor{blue}{$\uparrow$ 11.0}  & \textcolor{blue}{$\uparrow$ 0.8}   &                                   & \textcolor{red}{$\downarrow$ 2.6}   & \textcolor{red}{$\times$ 14.00}      & \textcolor{red}{$\times$ 5.81}      \\  \cmidrule{2-9}
& \multirow{2}{*}{YOLOPv3 \cite{yolopv3}}                  & 96.9                               & 84.3                               & 93.2                               & –                                 & 28.0                               & 30.2M                                 & 35.38G                               \\
&                                                          & \textcolor{blue}{$\uparrow$ 11.3}  & \textcolor{blue}{$\uparrow$ 12.0}  & \textcolor{blue}{$\uparrow$ 0.8}   &                                   & \textcolor{red}{$\downarrow$ 1.8}   & \textcolor{red}{$\times$ 12.85}      & \textcolor{red}{$\times$ 4.58}      \\ \bottomrule
\end{tabular}
\end{table*}

\subsection{Main results}

\begin{table*}[]
\centering
\caption{Performance and computational cost of TriLiteNet across different configurations}
\label{table:config_re}
\begin{tabular}{lccccccc}
\toprule
\multirow{2}{*}{\textbf{Config}} & \multicolumn{2}{c}{\textbf{{\footnotesize Vehicle Detection}}} & \textbf{{\footnotesize Drivable Are Segmentation}} & \multicolumn{2}{c}{\textbf{{\footnotesize Lane Line Segmentation}}} & \multirow{2}{*}{\textbf{Params (M)}} & \multirow{2}{*}{\textbf{GFLOPs}} \\ \cmidrule(lr){2-3}\cmidrule(lr){4-4}\cmidrule(lr){5-6}
                                 & \textbf{Recall (\%)}  & \textbf{mAP@0.5 (\%)}  & \textbf{mIoU (\%)}             & \textbf{Acc (\%)}   & \textbf{IoU (\%)}   &                                      &                                  \\ \midrule
TriLiteNet$_{tiny}$              & 76.5                  & 49.6                   & 88.5                           & 75.6                     & 24.2                & 0.15                                 & 0.55                             \\
TriLiteNet$_{small}$                       & 81.6                  & 63.2                   & 91.0                           & 79.5                     & 27.6                & 0.59                                 & 1.99                             \\ 

TriLiteNet$_{base}$                                        & 85.6                               & 72.3                               & 92.4                               & 82.3                              & 29.8                                & 2.35 & 7.72 \\ \bottomrule
\end{tabular}%
\end{table*}

\begin{table*}[]

\centering
\caption{Evaluation between multi-tasking learning versus single-task learning approach. \textbf{Det} represents the task of Vehicle Detection, whereas \textbf{DA} and \textbf{LL} correspond to Drivable Area Segmentation and Lane Line Segmentation, respectively.}
\label{tab:single2multi}
\begin{tabular}{cccccccccc}
\toprule
\multirow{2}{*}{\textbf{Det}} & \multirow{2}{*}{\textbf{DA}} & \multirow{2}{*}{\textbf{LL}} & \multicolumn{2}{c}{\textbf{{\footnotesize Vehicle Detection}}} & \textbf{{\footnotesize Drivable Are Segmentation}} & \multicolumn{2}{c}{\textbf{{\footnotesize Lane Line Segmentation}}} & \multirow{2}{*}{\textbf{Params (M)}} & \multirow{2}{*}{\textbf{GFLOPs}} \\  \cmidrule(lr){4-5}\cmidrule(lr){6-6}\cmidrule(lr){7-8}
                              &                              &                              & \textbf{Recall (\%)}   & \textbf{mAP@0.5 (\%)}  & \textbf{mIoU (\%)}             & \textbf{Acc (\%)}      & \textbf{IoU(\%)}      &                                      &                                  \\ \midrule
\checkmark                    &                              &                              & 81.3                   & \textbf{63.3}          & –                    & –           & –            & 0.47                                 & 1.18G                            \\
                              & \checkmark                   &                              & –             & –             & 90.8                           & –            &  –            & 0.18                                 & 1.47G                            \\
                              &                              & \checkmark                   &  –             &  –             &  –                     & 75.8                   & \textbf{29.3}         & 0.18                                 & 1.47G                            \\ \midrule
\checkmark                    & \checkmark                   & \checkmark                   & \textbf{81.6}          & 63.2                   & \textbf{91.0}                  & \textbf{79.5}          & 27.6                  & 0.59                                 & 1.99G                            \\ \bottomrule
\end{tabular}%
\end{table*}

\subsubsection{Computation Cost}

We conducted a computational cost comparison of the TriLiteNet model with other multi-task models \cite{yolop,yolopv2,yolopv3,hybridnets,YOLOPX,ayolom}. In addition to Parameters and FLOPs, the inference speed (FPS) was evaluated based on inference time at batch sizes of 1, 8, and 32 (excluding preprocessing and postprocessing times). The FPS metrics we reproduce for the models: YOLOP \footnote{\url{https://github.com/hustvl/YOLOP}}, YOLOPv2\footnote{\url{https://github.com/CAIC-AD/YOLOPv2}}, YOLOPv3\footnote{\url{https://github.com/jiaoZ7688/YOLOPv3}}, Hybridnets \footnote{\url{https://github.com/datvuthanh/HybridNets}}, YOLOPX\footnote{\url{https://github.com/jiaoZ7688/YOLOPX}}, A-YOLOM\footnote{\url{https://github.com/JiayuanWang-JW/YOLOv8-multi-task}} on our experimental devices are calculated 100 times and averaged to ensure fairness. The results in Table \ref{tab:cost} demonstrate that TriLiteNet configurations outperform other models in terms of computational efficiency. Notably, TriLiteNet$_{tiny}$ achieved the highest FPS, with 185, 1340, and 3397 at batch sizes of 1, 8, and 32, respectively, significantly exceeding models such as YOLOPX (49, 199, 262) and HybridNets (8, 53, 121). With only 0.15M parameters and 0.55G FLOPs, TriLiteNet$_{tiny}$ is particularly suitable for deployment in resource-constrained systems. Meanwhile, largest config, TriLiteNet$_{base}$ achieves FPS of 151, 1081, and 1641 at batch sizes 1, 8, and 32, with 2.35M parameters and 7.72G FLOPs. These findings highlight that TriLiteNet configurations not only reduce computational costs significantly but also fulfill real-time performance requirements, making them ideal solutions for multi-task applications in resource-constrained environments, particularly in autonomous driving systems.

\subsubsection{Comparison with State-of-the-Art Methods}

To demonstrate the superior balance between performance and computational cost of TriLiteNet, we compare it with other multi-task models \cite{yolop, yolopv2, yolopv3,hybridnets,ayolom,YOLOPX}. For a more comprehensive evaluation, we also compare it with single-task models: Vehicle Detection \cite{yolov5, faster, yolov8, multinet}, Drivable Are Segmentation \cite{GCNet, dnlnet, pspnet}, and Lane Line Segmentation \cite{enet, enet-sad, scnn}. In this section, we select the TriLiteNet$_{base}$ model for comparison, as shown in Table \ref{tab:comparison}. The results demonstrate that TriLiteNet$_{base}$ achieves impressive performance, particularly in segmentation tasks, with mIoU of 92.4\% for drivable area segmentation and Acc of 82.3\% for lane line segmentation, along with IoU of 29.8\%. These results validate the effectiveness of TriLiteNet$_{base}$'s segmentation-focused architecture. While performance of TriLiteNet$_{base}$ in vehicle detection (Recall: 85.6\%, mAP@0.5: 72.3\%) is lower than some models, this is an expected result due to its design prioritizing segmentation tasks. The trade-off between segmentation and detection is a key consideration in multi-task models, as optimizing for one task may impact the performance of the other. In the TriLiteNet$_{base}$, the focus on segmentation tasks, such as drivable area and lane line detection, allows the model to achieve high accuracy in these areas, which are critical for autonomous driving systems. However, this prioritization may slightly reduce the model's ability to detect smaller or overlapping vehicles, as the shared encoder and feature extraction process are optimized for segmentation. Notably, the design trade-offs allow TriLiteNet$_{base}$ to achieve competitive performance in segmentation tasks compared to other multi-task models.

Compared to single-task models, TriLiteNet$_{base}$ demonstrates its ability to manage multiple tasks while maintaining competitive performance. For vehicle detection, TriLiteNet$_{base}$ still outperforms several other models, such as Faster RCNN \cite{faster}, MultiNet \cite{multinet}, and R-CNNP (DET) \cite{yolop}, in terms of Recall and mAP, and even surpasses YOLOv8 (n) \cite{yolov8} in Recall, indicating its strong detection ability despite managing additional segmentation tasks. In Drivable Area Segmentation, TriLiteNet$_{base}$ surpasses models like GCNet (82.07\% mIoU) and DNLNet (84.35\% mIoU), and performs closely to PSPNet (89.6\% mIoU), a model specifically designed for this task. Similarly, TriLiteNet$_{base}$ achieves 29.8\% IoU for Lane Line Segmentation, significantly outperforming ENet-SAD (16.02\%) and SCNN (15.84\%) while providing additional multi-tasking capabilities. These comparisons underscore TriLiteNet$_{base}$'s versatility, as it balances segmentation and detection performance effectively.

With IoU of 29.8\% along with Acc of 82.3\% for Lane Line Segmentation, TriLiteNet$_{base}$ outperforms many prior multi-task models with lower computation costs. In Drivable Area Segmentation, TriLiteNet$_{base}$ achieves 92.4\% mIoU, only slightly lower than models such as YOLOPX, YOLOPv2, and YOLOPv3 (all achieving 93.2\% mIoU). However, these models require significantly more parameters (over 30 million compared to 2.35 million for TriLiteNet$_{base}$) and much higher computational cost. When compared to A-YOLOM (n), a model with a similar computational cost (4.43 million parameters and 6.66 GFLOPs), TriLiteNet$_{base}$ demonstrates superior performance in segmentation tasks. Specifically, in Drivable Area Segmentation, TriLiteNet$_{base}$ achieves a 1.9\% higher mIoU, while in Lane Line Segmentation, it achieves 1.0\% higher accuracy (Acc) and 1.6\% higher IoU. Although the mAP of TriLiteNet$_{base}$ is 5.7\% lower than that of A-YOLOM (n), due to the latter's design focused on object detection. This demonstrates that TriLiteNet$_{base}$'s segmentation-focused design achieves competitive performance with a fraction of the computational resources.


In summary, while TriLiteNet does not achieve the highest performance in vehicle detection due to its segmentation-focused design, it excels in segmentation tasks. It maintains competitive overall performance with significantly lower computational resources. Compared to both multi-task models and single-task models, TriLiteNet demonstrates a balanced performance-cost trade-off, making it an efficient solution for embedded systems and real-time applications, particularly in scenarios with limited computational resources.

\subsubsection{Compare TriLiteNet's configurations}

In Table \ref{table:config_re}, we compare the different configurations of the TriLiteNet model. The results demonstrate that computational efficiency improves proportionally as model complexity increases (as described in Table \ref{table:config}). Specifically, the TriLiteNet$_{base}$ configuration achieves superior performance across all tasks. However, smaller configurations, such as TriLiteNet$_{small}$ (0.59M parameters, 1.99 GFLOPs) and TriLiteNet$_{tiny}$ (0.15M parameters, 0.55 GFLOPs), still deliver impressive results across all three tasks despite their significantly reduced complexity. These results indicate that TriLiteNet$_{tiny}$ and TriLiteNet$_{small}$ can effectively perform all tasks with stable performance while requiring minimal computational costs. However, with the achieved performance, the smaller configurations can meet the requirements of driving systems in fixed or simple environments and are fully suitable for embedded systems that demand resource optimization. In contrast, TriLiteNet$_{base}$ (2.35M parameters, 7.72 GFLOPs) is an ideal choice for applications requiring higher performance, offering an optimal balance between computational efficiency and accuracy.

\begin{table*}[]
\caption{Performance of TriLiteNet across different photometric conditions in panoptic driving perception}
\label{tab:panoptic}
\centering
\begin{tabular}{lllccccc}
\toprule
\multicolumn{2}{l}{\multirow{2}{*}{\textbf{Photometric Conditions}}}               & \multirow{2}{*}{\footnotesize \textbf{Number}} & \multicolumn{2}{c}{\footnotesize \textbf{Vehicle Detection}}             & \footnotesize \textbf{Drivable Are Segmentation} & \multicolumn{2}{c}{\footnotesize \textbf{Lane Line Segmentation}} \\ \cmidrule(lr){4-5} \cmidrule(lr){6-6} \cmidrule(lr){7-8}
\multicolumn{2}{l}{}                                &                                                & \footnotesize \textbf{Recall (\%)} & \footnotesize \textbf{mAP@0.5 (\%)} & \footnotesize \textbf{mIoU (\%)}                 & \footnotesize \textbf{Acc (\%)} & \footnotesize \textbf{IoU (\%)} \\ \midrule
\multirow{4}{*}{Times of the day}   & Daytime       & 5,258                                          & 80.4                               & 63.7                                & 91.2                                             & 80.0                            & 27.9                            \\
                                    & Night         & 3,929                                          & 83.8                               & 62.1                                & 90.9                                             & 78.7                            & 27.2                            \\
                                    & Dawn/Dusk     & 778                                            & 80.3                               & 63.0                                & 90.8                                             & 79.7                            & 27.5                            \\
                                    & Undefined     & 35                                             & 86.8                               & 70.0                                & 84.6                                             & 72.3                            & 23.3                            \\ \midrule
\multirow{7}{*}{Weather conditions} & Clear         & 5,346                                          & 81.8                               & 62.4                                & 91.4                                             & 79.3                            & 27.5                            \\
                                    & Overcast      & 1,239                                          & 80.4                               & 64.4                                & 91.9                                             & 82.5                            & 28.8                            \\
                                    & Undefined     & 1,157                                          & 83.2                               & 67.5                                & 91.0                                             & 80.8                            & 28.3                            \\
                                    & Snowy         & 769                                            & 81.5                               & 61.2                                & 89.1                                             & 73.7                            & 24.7                            \\
                                    & Rainy         & 738                                            & 83.0                               & 63.0                                & 87.8                                             & 74.9                            & 25.5                            \\
                                    & Partly Cloudy & 738                                            & 78.7                               & 61.7                                & 91.4                                             & 80.6                            & 28.3                            \\
                                    & Foggy         & 13                                             & 82.5                               & 60.5                                & 88.7                                             & 69.7                            & 22.7                            \\ \midrule
\multirow{7}{*}{Scene types}        & City street   & 6,112                                          & 83.0                               & 64.4                                & 90.9                                             & 79.8                            & 27.8                            \\
                                    & Highway       & 2,499                                          & 76.7                               & 56.6                                & 91.4                                             & 78.8                            & 27.4                            \\
                                    & Residential   & 1,253                                          & 82.5                               & 67.9                                & 90.9                                             & 80.2                            & 27.5                            \\
                                    & Undefined     & 53                                             & 86.7                               & 59.4                                & 85.5                                             & 72.4                            & 24.3                            \\
                                    & Parking lot   & 49                                             & 875                                & 60.8                                & 87.3                                             & 66.0                            & 18.6                            \\
                                    & Tunnel        & 27                                             & 88.2                               & 73.1                                & 89.5                                             & 70.2                            & 22.4                            \\
                                    & Gas station   & 7                                              & 82.5                               & 50.7                                & 76.2                                             & 59.9                            & 12.3                            \\ \midrule
\multicolumn{2}{l}{Total}                           & 10,000                                         & 81.6                               & 63.2                                & 91.0                                             & 79.5                            & 27.6                            \\ \bottomrule
\end{tabular}%
\end{table*}

\begin{table}[]
\centering
\caption{Results of the TriLiteNet$_{base}$ model under different precision levels.}
\label{tab:quant}
\begin{tabular}{lccccc}
\toprule
\multirow{3}{*}{} & \multicolumn{2}{c}{\multirow{2}{*}{\textbf{Vehicle Detection}}} & \textbf{Drivable Area}                    & \multicolumn{2}{c}{\textbf{Lane Line}}    \\ 
                                    & \multicolumn{2}{c}{}                                            & \multicolumn{1}{l}{\textbf{Segmentation}} & \multicolumn{2}{l}{\textbf{Segmentation}} \\
                                    \cmidrule(lr){2-3} \cmidrule(lr){4-4}\cmidrule(lr){5-6}
                                    & \textbf{Recall}                & \textbf{mAP@0.5}               & \textbf{mIoU}                             & \textbf{Acc}         & \textbf{IoU}       \\ \midrule
FP32                                & 81.6\%                  & 63.2\%                   & 91.0\%                           & 79.5\%                     & 27.6\%             \\
FP16                                & 81.6\%                  & 63.2\%                   & 91.0\%                           & 79.5\%                     & 27.6\%             \\
INT8                                & 81.2\%                         & 61.7\%                         & 90.5\%                                    & 78.2\%               & 27.1\%             \\ \bottomrule
\end{tabular}%
\end{table}

\subsection{Ablation study}

We conducted all ablation experiments to demonstrate the effectiveness of the TriLiteNet model in various aspects. In addition, in this section, we report results using the TriLiteNet$_{small}$ configuration.

\subsubsection{Multi-task versus single-task}

To evaluate the efficiency of our end-to-end model, we compared the performance of the multi-task model with single-task models. Table \ref{tab:single2multi} demonstrates detailed performance metrics and computational costs for each design, including single-task and multi-task configurations. The results demonstrate that our multi-task model achieves performance levels very close to those of the single-task models on individual tasks. Notably, in some cases, multi-task training not only maintains effectiveness but also improves results, achieving higher performance than single-task training. Specifically, for vehicle detection, the multi-task model achieves mAP@0.5 of 63.2\%, which is very close to the 63.3\% obtained through single-task training. However, it shows an improvement in Recall, reaching 81.6\% compared to 81.3\% of the single-task model. For drivable area segmentation, the multi-task model reaches mIoU of 91.0\%, surpassing the single-task model's mIoU of 90.8\%. In the lane line segmentation task, while the multi-task model achieves IoU of 27.6\%, which is lower than the single-task IoU of 29.3\%, it boasts an Acc of 79.5\%, 3.7\% higher than the Acc of the single-task model.

In summary, our study shows that adopting a multi-task model saves computational resources and, in some cases, enhances performance compared to single-task models. Thanks to its optimized design, our multi-task model offers an effective solution for real-world applications where both high performance and low computational cost are essential.

\subsubsection{Panoptic driving perception results}

We evaluated the performance of the TriLiteNet model on the BDD100K dataset, which provides a diverse set of scenarios categorized by time of day, weather conditions, and scene types. This comprehensive dataset allows us to assess the model's robustness and adaptability across different environmental conditions. The results presented in Table \ref{tab:panoptic} indicate that TriLiteNet maintains stable performance across various driving conditions, including varying lighting, weather, and scene complexities. These findings highlight the model's potential for real-world deployment in autonomous driving systems, where consistency across dynamic and challenging environments is crucial.

\subsubsection{Quantization}

We evaluate the TriLiteNet model at different precision levels, including floating point 32-bit (FP32), floating point 16-bit (FP16), and integer 8-bit (INT8). To quantize the model to INT8, we apply the post-training static quantization method, with weights calibrated using 500 samples from the validation set. The results shown in Table \ref{tab:quant} indicate that FP16 inference maintains the exact accuracy of the model compared to FP32. Meanwhile, when using INT8, there is a slight performance degradation, but the reduction is not significant. This suggests that quantization is a reasonable approach to significantly reducing computational costs, especially when deploying on embedded devices or real-time inference systems.

This is particularly important for systems utilizing GPUs or hardware that supports quantization, such as NVIDIA TensorRT, which optimizes inference speed while maintaining high performance. These findings confirm that TriLiteNet can effectively operate at different precision levels, offering high flexibility for deployment across various hardware platforms.

\subsubsection{The ablations in different experimental settings}

We conducted an ablation study on various design choices in the TriLiteNet model and the EMA method to evaluate the role of each factor in improving performance. The results in Table \ref{tab:ab} demonstrate that components such as PCAA (Partial Class Activation Attention) \cite{pcaa}, LitePAN proposed, SPP (Spatial Pyramid Pooling)\cite{spp}, and EMA (Exponential Moving Average) \cite{ema} significantly contribute to the overall performance of TriLiteNet. Specifically, the baseline model initially shows stable performance. However, as each component is incrementally added, the model’s performance improves markedly. Integrating PCAA enhances image segmentation capabilities, improving performance in drivable area and lane line segmentation tasks. Then, LitePAN and SPP enhance the ability to detect small objects in highly complex images. Although combining these components may increase computational cost, careful optimization ensures the model maintains a reasonable computational. Notably, applying EMA during training achieves optimal performance. EMA not only accelerates and stabilizes convergence but also maintains high performance across detection and segmentation tasks, yielding better results without significantly increasing the parameter or computational cost. 

This comparison highlights the essential role of each design component in enhancing TriLiteNet’s performance. Each component optimizes different aspects of the model, and when combined, they enable TriLiteNet to achieve ideal multi-tasking results while maintaining low computational costs.

\begin{table*}[]
\centering
\setlength{\tabcolsep}{9.5pt}
\caption{Ablation experiments results. }
\label{tab:ab}
\begin{tabular}{lccccccc}
\toprule
\multirow{2}{*}{\textbf{Method}} & \multicolumn{2}{c}{\textbf{Vechicle Detection}}                                                                                                                                   & \textbf{Drivable Are Segmentation}                                                          & \multicolumn{2}{c}{\textbf{Lane Line Segmentation}}                                                                                                                                     & \multirow{2}{*}{\textbf{Params (M)}}                                            & \multirow{2}{*}{\textbf{FLOPs}}                                                  \\ \cmidrule(lr){2-3}\cmidrule(lr){4-4}\cmidrule(lr){5-6}
                                 & \textbf{Recall (\%)}                                                                    & \textbf{mAP@0.5 (\%)}                                                                   & \textbf{mIoU (\%)}                                                                      & \textbf{Acc (\%)}                                                                       & \textbf{IoU (\%)}                                                                        &                                                                                 &                                                                                  \\ \midrule
Baseline                         & 78.4                                                                                    & 58.3                                                                                    & 88.8                                                                                    & 77.9                                                                                    & 25.9                                                                                     & \textbf{0.45}                                                                   & \textbf{1.77G}                                                                   \\ \midrule
+ PCAA                            & \begin{tabular}[c]{@{}c@{}}78.7\\ \textcolor{blue}{$\uparrow$ 0.3}\end{tabular}         & \begin{tabular}[c]{@{}c@{}}58.5\\ \textcolor{blue}{$\uparrow$ 0.2}\end{tabular}         & \begin{tabular}[c]{@{}c@{}}90.2\\ \textcolor{blue}{$\uparrow$ 1.4}\end{tabular}         & \begin{tabular}[c]{@{}c@{}}78.5\\ \textcolor{blue}{$\uparrow$ 0.6}\end{tabular}         & \begin{tabular}[c]{@{}c@{}}27.0\\ \textcolor{blue}{$\uparrow$ 1.1}\end{tabular}          & \begin{tabular}[c]{@{}c@{}}0.51\\ \textcolor{red}{$\uparrow$ 0.06}\end{tabular} & \begin{tabular}[c]{@{}c@{}}1.93G\\ \textcolor{red}{$\uparrow$ 0.16}\end{tabular} \\ \midrule
+ LitePAN                         & \begin{tabular}[c]{@{}c@{}}80.1\\ \textcolor{blue}{$\uparrow$ 1.7}\end{tabular}         & \begin{tabular}[c]{@{}c@{}}60.6\\ \textcolor{blue}{$\uparrow$ 2.3}\end{tabular}         & \begin{tabular}[c]{@{}c@{}}90.4\\ \textcolor{blue}{$\uparrow$ 1.6}\end{tabular}         & \begin{tabular}[c]{@{}c@{}}79.0\\ \textcolor{blue}{$\uparrow$ 1.1}\end{tabular}         & \begin{tabular}[c]{@{}c@{}}\textbf{27.7}\\ \textcolor{blue}{$\uparrow$ 1.8}\end{tabular} & \begin{tabular}[c]{@{}c@{}}0.55\\ \textcolor{red}{$\uparrow$ 0.10}\end{tabular} & \begin{tabular}[c]{@{}c@{}}1.98G\\ \textcolor{red}{$\uparrow$ 0.21}\end{tabular} \\ \midrule
+ SPP                             & \begin{tabular}[c]{@{}c@{}}81.3\\ \textcolor{blue}{$\uparrow$ 2.9}\end{tabular}         & \begin{tabular}[c]{@{}c@{}}62.1\\ \textcolor{blue}{$\uparrow$ 3.8}\end{tabular}         & \begin{tabular}[c]{@{}c@{}}90.5\\ \textcolor{blue}{$\uparrow$ 1.7}\end{tabular}         & \begin{tabular}[c]{@{}c@{}}79.2\\ \textcolor{blue}{$\uparrow$ 1.3}\end{tabular}         & \begin{tabular}[c]{@{}c@{}}\textbf{27.7}\\ \textcolor{blue}{$\uparrow$ 1.8}\end{tabular}          & \begin{tabular}[c]{@{}c@{}}0.59\\ \textcolor{red}{$\uparrow$ 0.14}\end{tabular} & \begin{tabular}[c]{@{}c@{}}1.99G\\ \textcolor{red}{$\uparrow$ 0.22}\end{tabular} \\ \midrule
+ EMA                             & \begin{tabular}[c]{@{}c@{}}\textbf{81.6}\\ \textcolor{blue}{$\uparrow$ 3.2}\end{tabular} & \begin{tabular}[c]{@{}c@{}}\textbf{63.2}\\ \textcolor{blue}{$\uparrow$ 4.9}\end{tabular} & \begin{tabular}[c]{@{}c@{}}\textbf{91.0}\\ \textcolor{blue}{$\uparrow$ 2.2}\end{tabular} & \begin{tabular}[c]{@{}c@{}}\textbf{79.5}\\ \textcolor{blue}{$\uparrow$ 1.6}\end{tabular} & \begin{tabular}[c]{@{}c@{}}27.6\\ \textcolor{blue}{$\uparrow$ 1.7}\end{tabular}           & \begin{tabular}[c]{@{}c@{}}0.59\\ \textcolor{red}{$\uparrow$ 0.14}\end{tabular} & \begin{tabular}[c]{@{}c@{}}1.99G\\ \textcolor{red}{$\uparrow$ 0.22}\end{tabular} \\ \bottomrule
\end{tabular}%
\end{table*}

\begin{table}[!t]
\centering
\caption{Latency and power consumption benchmarking on embedded dvevices}
\label{tab:latency}
\begin{tabular}{clcc}
\toprule
& {\textbf{Configuration}} & {\textbf{Jetson Xavier}} & {\textbf{Jetson TX2}} \\
\midrule
\addlinespace[0.5em]
\multirow{3}{*}{\begin{tabular}[c]{@{}c@{}}\textbf{Latency} \textbf{\footnotesize{(ms)}}\end{tabular}}
& TriLiteNet$_{tiny}$     & 8.962$_{\pm 0.02}$  & 23.420$_{\pm 0.02}$  \\
\addlinespace[0.8em]
& TriLiteNet$_{small}$    & 16.791$_{\pm 0.03}$ & 39.100$_{\pm 0.04}$  \\
\addlinespace[0.8em]
& TriLiteNet$_{base}$     & 25.822$_{\pm 0.05}$  & 48.420$_{\pm 0.02}$  \\
\addlinespace[0.5em]
\midrule
\addlinespace[0.5em]
\multirow{3}{*}{\begin{tabular}[c]{@{}c@{}}\textbf{Power} \textbf{\footnotesize{(Watt)}}\end{tabular}} 
& TriLiteNet$_{tiny}$     & 11.908$_{\pm 2.23}$ & 3.595$_{\pm 1.91}$   \\
\addlinespace[0.8em]
& TriLiteNet$_{small}$    & 17.914$_{\pm 4.90}$ & 4.190$_{\pm 1.77}$   \\
\addlinespace[0.8em]
& TriLiteNet$_{base}$    & 22.824$_{\pm 3.15}$ & 4.590$_{\pm 1.57}$   \\
\addlinespace[0.4em]
\bottomrule
\end{tabular}
\end{table}

\subsection{Deployment}

To evaluate inference latency and power consumption, we deployed the configurations of the TriLiteNet model on embedded devices, including Jetson Xavier and Jetson TX2. The results, presented in Table \ref{tab:latency}, were obtained using TensorRT-FP16 and demonstrate that TriLiteNet can achieve low-latency inference on embedded devices. TriLiteNet$_{tiny}$ achieved outstanding inference latency, with only 8.962 ms on Jetson Xavier and 23.42ms on Jetson TX2, making it an ideal choice for real-time applications on resource-constrained devices. Although the configurations TriLiteNet$_{small}$ and TriLiteNet$_{base}$ exhibit higher latency compared to TriLiteNet$_{tiny}$, the results confirm that all three configurations maintain low-latency inference on the tested embedded platforms. Regarding power consumption, the data indicate that TriLiteNet$_{tiny}$ consumes an average of 11.908W on Jetson Xavier and 3.595W on Jetson TX2. In contrast, the TriLiteNet$_{base}$ configuration requires higher power, consuming 22.824W on Jetson Xavier and 4.590W on Jetson TX2. To ensure the reliability of these measurements, each configuration was tested independently five times. The average values and standard deviations were calculated to accurately reflect the model's real-world performance, ensuring the results are trustworthy.

These results underscore the proposed approach's practical applicability. All TriLiteNet configurations exhibit fast inference and low energy consumption on embedded devices like the Jetson Xavier and Jetson TX2, demonstrating their ability for real-time applications, especially in scenarios requiring a balance between responsiveness and energy efficiency.

\subsection{Visualization}

\begin{figure*}[!t]
    \centering
    \includegraphics[width=\textwidth]{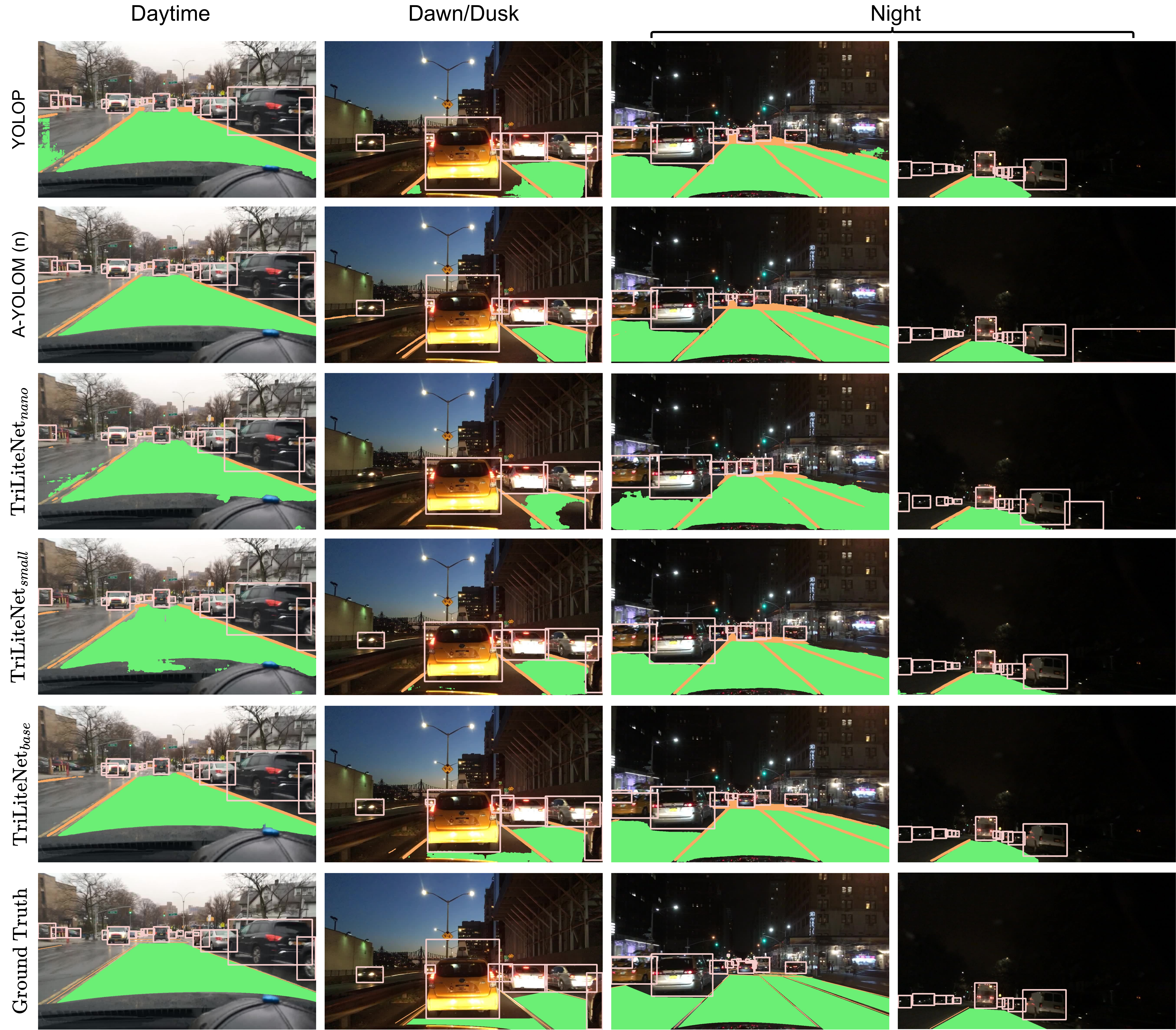}
    \caption{Visual comparison results of models under different time types: \textit{Daytime, Dawn/Dusk, and Night}.}
    \label{fig:vs_time}

\end{figure*}
\begin{figure*}[ht]
    \centering
    \includegraphics[width=\textwidth]{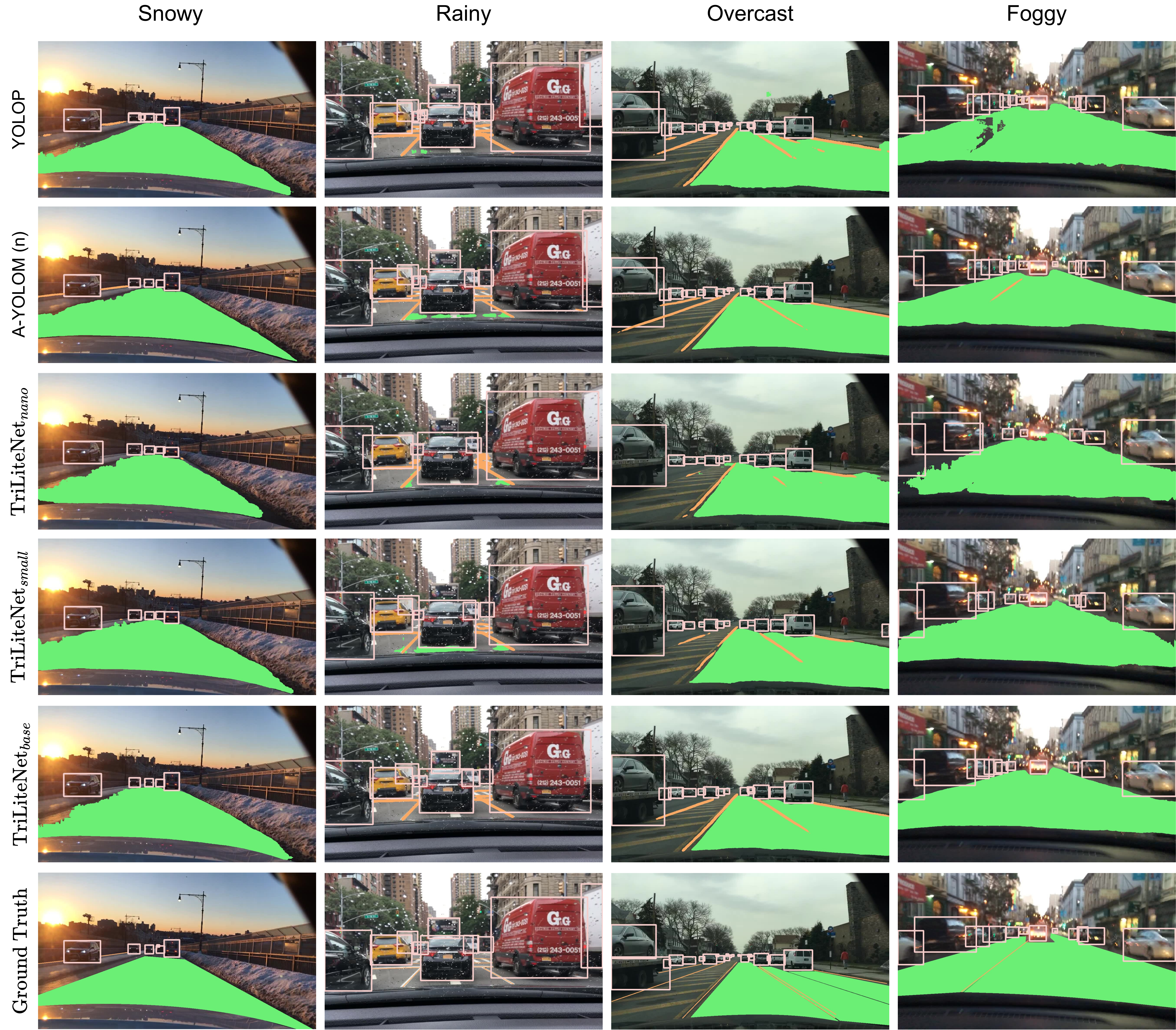}
    \caption{Visual comparison results of models under different weather conditions: \textit{Snowy, Rainy, Overcast, and Foggy}.}
    \label{fig:vs_weather}

\end{figure*}
\begin{figure*}[ht]
    \centering
    \includegraphics[width=\textwidth]{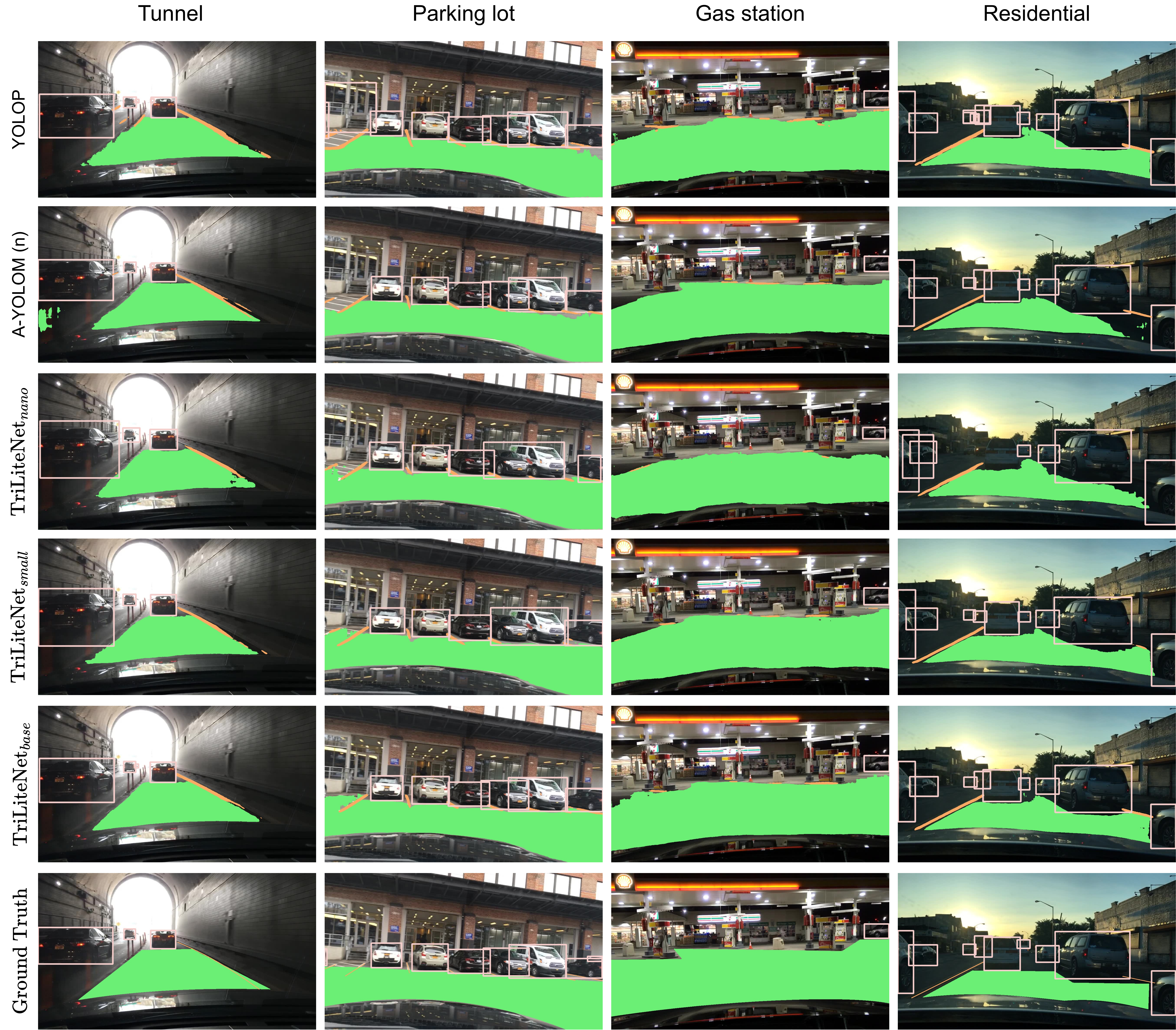}
    \caption{Visual comparison results of models under different scene conditions: \textit{Tunnel, Parking lot, Gas station,  Residential}.}
    \label{fig:vs_scene}

\end{figure*}

For a comprehensive evaluation, in addition to thoroughly comparing the configurations of TriLiteNet, we also compare the qualitative results of TriLiteNet with YOLOP and A-YOLOM (n). These two models have a computational cost similar to TriLiteNet, making them appropriate baselines for comparison when performing inference on the validation set of the BDD100K dataset. The visualization results, categorized by different weather conditions, scene types, and times of the day, are presented in Figures \ref{fig:vs_time},\ref{fig:vs_weather}, and \ref{fig:vs_scene}:
(1) Figure \ref{fig:vs_time} illustrates the results across different times of the day, including daytime, dawn/dusk, and night.
(2) Figure \ref{fig:vs_weather} presents the results under various weather conditions, including snowy, rainy, overcast, and foggy.
(3) Figure \ref{fig:vs_scene} showcases the results for different scene types, such as tunnels, parking lots, gas stations, and residential areas. Compared to YOLOP and A-YOLOM (n), the comparative visualization results indicate that TriLiteNet$_{base}$ demonstrates remarkable performance across all three tasks under varying environmental conditions. In addition to its strong segmentation performance, TriLiteNet$_{base}$ also delivers notable results in object detection despite this task not being the primary focus of optimization. Meanwhile, although TriLiteNet$_{tiny}$ and TriLiteNet$_{small}$ exhibit less prominent results, their lightweight design ensures promising performance even under challenging conditions. These findings highlight the potential of all configurations, particularly in scenarios where resource constraints are critical.

\section{\uppercase{Conclusions AND FUTURE WORK}}
\label{sec:conclusion}

In this study, we presented TriLiteNet, a lightweight multi-task perception model designed explicitly for resource-constrained environments in autonomous driving. By adopting a shared encoder-decoder architecture and three distinct decoders, TriLiteNet performs efficiently across three critical tasks: vehicle detection, drivable area segmentation, and lane line segmentation. Our experimental results on the BDD100K dataset demonstrate the competitiveness of TriLiteNet, achieving high metrics across all tasks while maintaining low computational costs. Furthermore, the three model configurations—TriLiteNet$_{tiny}$, TriLiteNet$_{small}$, and TriLiteNet$_{base}$—enable flexibility in deployment, catering to various levels of resource availability. Deployment experiments on embedded devices, including Jetson Xavier and TX2, confirmed the model’s low latency and energy efficiency, solidifying its suitability for real-world applications.

TriLiteNet’s ability to balance computational demands and performance makes it a robust solution for autonomous driving systems. However, challenges remain in real-world deployments, such as dynamic traffic conditions, varying workloads, and changing environmental factors. Addressing these challenges in future research will further enhance the model's robustness and practical applicability. This study underscores the importance of optimizing multi-task models for practical deployment and paves the way for future research in intelligent vehicles' scalable, resource-efficient perception systems.

\section*{Acknowledgment}
This research was supported by The VNUHCM-University of Information Technology’s Scientific Research Support Fund.

\bibliographystyle{unsrt}
\bibliography{ref}

\end{document}